\documentclass[manuscript,screen]{acmart}

\usepackage{times} 
\usepackage{soul}
\usepackage{url}
\usepackage{graphicx}
\usepackage{amsmath}
\usepackage{amsthm}
\usepackage{booktabs}
\usepackage{algorithm}
\usepackage{algorithmic}
\usepackage{wrapfig}
\usepackage{float}
\usepackage{multirow} 
\usepackage{subcaption}
\usepackage{rotating}
\usepackage{ragged2e}

\AtBeginDocument{%
  \providecommand\BibTeX{{%
    \normalfont B\kern-0.5em{\scshape i\kern-0.25em b}\kern-0.8em\TeX}}}

\setcopyright{acmcopyright}
\copyrightyear{2018}
\acmYear{2018}
\acmDOI{XXXXXXX.XXXXXXX}

\acmConference[Conference acronym 'XX]{Make sure to enter the correct
  conference title from your rights confirmation emai}{June 03--05,
  2018}{Woodstock, NY}
\acmPrice{15.00}
\acmISBN{978-1-4503-XXXX-X/18/06}

\begin{document}

\title{Surrogate Lagrangian Relaxation: A Path To Retrain-free Deep Neural Network Pruning}

\author{Shanglin Zhou}
\email{shanglin.zhou@uconn.edu}
\orcid{0000-0002-6409-7716}
\affiliation{%
  \institution{University of Connecticut}
  \streetaddress{371 Fairfield Way}
  \city{Storrs}
  \state{Connecticut}
  \country{USA}
  \postcode{06269}
}

\author{Mikhail A. Bragin}
\orcid{0000-0002-7783-9053}
\affiliation{%
  \institution{University of Connecticut}
  \city{Storrs}
  \country{USA}
}

\author{Lynn Pepin}
\affiliation{%
  \institution{University of Connecticut}
  \city{Storrs}
  \country{USA}
}

\author{Deniz Gurevin}
\affiliation{%
  \institution{University of Connecticut}
  \city{Storrs}
  \country{USA}
}

\author{Fei Miao}
\affiliation{%
  \institution{University of Connecticut}
  \city{Storrs}
  \country{USA}
}

\author{Caiwen Ding}
\affiliation{%
  \institution{University of Connecticut}
  \city{Storrs}
  \country{USA}
}
\email{caiwen.ding@uconn.edu}

%
\renewcommand{\shortauthors}{Zhou, et al.}

\begin{abstract}
Network pruning is a widely used technique to reduce computation cost and model size for deep neural networks. However, the typical three-stage pipeline, i.e., training, pruning, and retraining (fine-tuning), significantly increases the overall training time. In this paper, we develop a systematic weight-pruning optimization approach based on Surrogate Lagrangian relaxation (SLR), which is tailored to overcome difficulties caused by the discrete nature of the weight-pruning problem. We further prove that our method ensures fast convergence of the model compression problem, and the convergence of the SLR is accelerated by using quadratic penalties. Model parameters obtained by SLR during the training phase are much closer to their optimal values as compared to those obtained by other state-of-the-art methods.
We evaluate our method on image classification tasks using CIFAR-10 and ImageNet with state-of-the-art multi-layer perceptrons (MLPs)-based networks such as MLP-Mixer, attention-based networks such as Swin Transformer, and convolutional neural networks-based models such as VGG-16, ResNet-18, ResNet-50 and ResNet-110, MobileNetV2. We also evaluate object detection and segmentation tasks on COCO, KITTI benchmark, and TuSimple lane detection dataset using a variety of models. Experimental results demonstrate that our SLR-based weight-pruning optimization approach achieves a higher compression rate than state-of-the-art methods under the same accuracy requirement and also can achieve higher accuracy under the same compression rate requirement. Under classification tasks, our SLR approach converges to the desired accuracy $3\times$ faster on both of the datasets. Under object detection and segmentation tasks, SLR also converges $2\times$ faster to the desired accuracy.
Further, our SLR achieves high model accuracy even at the hard-pruning stage without retraining, which reduces the traditional three-stage pruning into a two-stage process. Given a limited budget of retraining epochs, our approach quickly recovers the model's accuracy.
\end{abstract}




\maketitle

\section{Introduction}

Deep neural network (DNN)-based statistical models are increasingly demanding of computational and storage resources, with costs proportional to the model size (i.e., the number of parameters in a model). This resource consumption is especially an issue for embedded or IoT devices~\cite{krizhevsky2012imagenet,simonyan2014}. By reducing model size, one can decrease both storage costs and computation costs when evaluating a model. Various techniques exist for reducing model size while maintaining its performance, e.g., weight pruning, sparsity regularization, quantization, and clustering. These techniques are collectively known as \emph{model compression} ~\cite{dai2017,yang2016,molchanov2017variational,guo2016dynamic,tung2017fine,luo2017thinet,ding2017circnn,park2017weighted,He_2018_ECCV}.

These works leverage the observation that training a compact model from scratch is more difficult and less effective than retraining a pruned model~\cite{frankle2018lottery,liu2018rethinking}. Therefore, a typical three-stage pipeline has been used: training (large model), pruning, and retraining (also called ``fine-tuning"). The pruning process involves setting the redundant weights to zero while keeping the important weights to maintain performance. The retraining process is necessary since the model accuracy significantly decreases after hardpruning. However, this three-stage weight pruning approach substantially adds to the overall training cost. For example, although the state-of-the-art weight pruning methods achieve a very high compression rate while maintaining the prediction accuracy on many DNN architectures, the retraining process requires more time, e.g., 80 epochs for ResNet-18 on ImageNet, which is $70\%$ of the original training epochs using Alternate Direction Method of Multipliers (ADMM)~\cite{zhang2018systematic,ren2019admm}.

Given the pros and cons of the current weight pruning method, this paper aims to answer the following questions: 
Is there an optimization method that can achieve high model accuracy even at the hard-pruning stage and can significantly reduce retraining trails? Given a limited budget of retraining epochs, is there an optimization method that can rapidly recover model accuracy (much faster than the state-of-the-art methods)?

The primary obstacle in addressing these questions is the discrete nature of the model compression problems caused by ``cardinality" constraints, which ensure that a certain proportion of weights is pruned. In this paper, we develop a weight-pruning optimization approach based on recent Surrogate Lagrangian relaxation (SLR) \cite{Bragin2015SLR}, which overcomes all major convergence difficulties of standard Lagrangian relaxation. Within the SLR approach, Lagrangian multipliers converge to their optimal values much faster as compared to those within other methods (e.g., ADMM). 

We summarize our contributions/findings as:

\begin{itemize}

\item We adapt the SLR-based approach to overcome difficulties caused by the discrete nature of the weight-pruning problem while ensuring fast convergence. 

\item We use quadratic penalties to further accelerate the SLR convergence. The method possesses nice convergence properties inherited from the rapid reduction of constraint violations owing to quadratic penalties, and quadratic penalties ultimately lead to faster convergence. Also, unlike previous methods such as ADMM, the SLR guarantees convergence, thereby leading to unmatched performance compared to other methods. Therefore, model parameters obtained by SLR are much closer to their optimal values as compared to those obtained by other state-of-the-art methods.

\item We provide a convergence proof of the SLR method for weight pruning problems. Existing coordination-based weight pruning approaches do not converge when solving non-convex problems. Other coordination techniques (e.g., ADMM) are not designed to handle discrete variables and other types of non-convexities. 

\item Our proposed SLR-based model-compression method achieves high model accuracy even at the hard-pruning stage using our SLR-based weight-pruning optimization approach; given a limited budget of retraining epochs, our method quickly recovers the model accuracy. 

\end{itemize}

We conduct comprehensive experiments on various tasks and datasets to further prove the effectiveness of our proposed SLR-based model compression method. 
For classification tasks, we test our method on not only convolutional neural networks (CNN)-based models like VGG-16, ResNet-18, ResNet-50 ResNet-110, and MobileNetV2, but also on non-CNN based models such as MLP-Mixer, a multi-layer perceptron (MLPs)-based network, and Swin Transformer, an attention-based network. We also test and compare our SLR method with other state-of-the-art (SoTA) pruning methods on segmentation and detection tasks. Our experiments involve various dataset benchmarks like CIFAR-10, ImageNet, COCO, KITTI, and TuSimple. The results demonstrate that our proposed SLR method outperforms the state-of-the-art compression methods. Our method converges $3\times$ faster to the desired accuracy on both CIFAR-10 and ImageNet datasets under both CNN-based and non-CNN-based classification tasks, $2\times$ faster on COCO object detection tasks. Moreover, up to a $6\%$ accuracy gap can be achieved between SLR and SoTA at the hard-pruning stage under classification tasks, and a $44\%$ accuracy gap in object detection and segmentation tasks. 


\section{Related Research}
 
\subsection{Model Compression}

Given the increasing computational and storage demands of Deep Neural Networks (DNNs), model compression has become increasingly essential when we implement highly efficient deep learning applications in the real world. There are two common compression techniques, weight pruning, and weight quantization. As numerous researchers have investigated that some portion of weights in neural networks are redundant, weight pruning aims to remove these less important coefficient values and it achieves model compression while maintaining performance similar to the uncompressed model. Structured and non-structured (irregular) weight pruning are two mainstream methods. Weight quantization is another technique that reduces weight storage by decreasing the number of bits used to represent weights. 

In early work~\cite{han2015learning}, the researchers proposed an iterative irregular weight pruning method where most reductions are achieved in fully-connected layers, and the reduction achieved in convolutional layers can hardly achieve significant acceleration in GPUs. For weight storage, it reduces $9\times$ the number of parameters in AlexNet and $13\times$ in VGGNet-16. To address the limitation in irregular weight pruning, structured weight pruning methods were proposed by~\cite{wen2016learning}. It investigated structured sparsity at the levels of filters, channels, and filter shapes. However, the overall compression rate in structured pruning is limited compared to unstructured pruning. In AlexNet without accuracy degradation, the average weight pruning rate in convolutional layers is only $1.4\times$. The recent work \cite{he2017channel} achieved $2\times$ channel pruning with a $1\%$ accuracy degradation on VGGNet-16. Later, \cite{louizos2018learning} proposed a framework for $L_0$ norm regularization for neural networks, aiming to prune the network during training by selecting weights and setting them to exactly zero. \cite{frankle2018lottery} introduced The Lottery Ticket Hypothesis, which observes that a subnetwork of a randomly-initialized network can replace the original network with the same performance. 

In this work, our focus is on irregular pruning which can achieve much higher accuracy compared to structured pruning~\cite{wen2016learning} due to its flexibility in selecting weights.

\subsection{Alternating Direction Method of Multipliers}

The ADMM is an optimization algorithm that breaks optimization problems into smaller subproblems, each of which is then solved iteratively and more easily. The early studies of ADMM can be traced back to the 1970s, and a variety of statistical and machine learning problems that can be efficiently solved by using ADMM were discussed~\cite{Boyd2011DistributedOA}. Recently, weight pruning studies achieved a high compression rate and avoided significant accuracy loss by integrating the powerful ADMM. The successful applications with ADMM outperform prior approaches by applying dynamic penalties on all targeted weights. The algorithm can be applied to various schemes of both nonstructured pruning and structured pruning. \cite{zhang2018systematic} was the first work implementing an ADMM-based framework on DNN weight pruning, achieving $21\times$ irregular weight pruning with almost no accuracy loss in AlexNet. A pattern-based weight pruning approach was proposed with high efficiency specifically designed and optimized for mobile devices~\cite{niu2020patdnn}, it explored a fine-grained sparsity to maximize the utilization of devices with limited resources. \cite{li2020ss} improved the previous ADMM-based structured weight pruning framework by adopting a soft constraint-based formulation to achieve a higher compression rate and tune fewer hyperparameters. 


\section{Weight Pruning using Surrogate Lagrangian Relaxation (SLR)}

Consider a deep neural network (DNN) with $N$ layers indexed by $n \in 1,...,N$, where the weights in layer $n$ are denoted by ${{\bf W}}_{n}$. The objective is to minimize a loss function 
\begin{eqnarray}
\underset{{\mathbf{W}}_{n}}{\text{min}} \; \left\{f \big( {\bf{W}}_{n}\big)
\right\}
\end{eqnarray}
\noindent subject to constraints on the cardinality of weights within each layer $n$, where the number of nonzero weights should be less than or equal to the predefined number $l_n$. This constraint can be captured using an indicator function $g_{n}(\cdot)$ as:

\begin{eqnarray}
g_{n}({\bf{W}}_{n})=
\begin{cases}
 0 & \text {if } \mathrm{card}({\bf{W}}_{n})\le l_{n}, \; n = 1, \ldots, N \\
 +\infty & \text {otherwise} 
\end{cases}
\end{eqnarray}

In its entirety, the problem cannot be solved either analytically or by using stochastic gradient descent.  To enable the decomposition into smaller manageable subproblems, duplicate variables are introduced and the problem is equivalently rewritten as:
\begin{equation}
\underset{{\mathbf{W}}_{n},{\mathbf{Z}}_{n}}{\text{min}} \; \left\{f \big( {\bf{W}}_{n}\big) + \sum_{n=1}^{N} g_{n}({\bf{Z}}_{n})\right\},
\quad \text{subject to}~{\bf{W}}_{n}={\bf{Z}}_{n}, \; n = 1, \ldots, N
\end{equation}
Here the first term is a nonlinear smooth loss function and the second term is a non-differentiable ``cardinality" penalty term~\cite{zhang2018systematic}.
To solve the problem, constraints are first relaxed by introducing Lagrangian multipliers to decompose the resulting problem into manageable subproblems, which will be coordinated by the multipliers. The constraint violations are also penalized by using quadratic penalties to speed up convergence. The resulting \textit{Augmented} Lagrangian function ~\cite{Boyd2011DistributedOA,zhang2018systematic} of the above optimization problem is thus given by:
\begin{equation}
 L_{\rho} \big( {\bf{W}}_{n}, {\bf{Z}}_{n} , {\bf{\Lambda}}_{n}  \big) = f \big( {\bf{W}}_{n} \big) \label{relaxedproblem} + \sum_{n=1}^{N} g_{n}({\bf{Z}}_{n}) +\sum_{n=1}^{N} \mathrm{tr} [{\bf{\Lambda}}_{n}^T({\bf{W}}_{n}-{\bf{Z}}_{n}) ] + \sum_{n=1}^{N} \frac{\rho}{2} \| {\bf{W}}_{n}-{\bf{Z}}_{n} \|_{F}^{2}
\end{equation}
where ${\bf{\Lambda}}_{n}$ is a matrix of Lagrangian multipliers corresponding to constraints ${\bf{W}}_{n}={\bf{Z}}_{n}$, and has the same dimension as ${\bf{W}}_{n}$. The positive scalar $\rho$ is the penalty coefficient, $\mathrm{tr}(\cdot)$ denotes the trace, $ \| \cdot \|_{F}^{2}$ denotes the Frobenius norm.  
    
In the following, we are motivated by decomposability enabled by SLR~\cite{Bragin2015SLR}, which overcame all major difficulties of standard Lagrangian Relaxation, significantly reducing zigzagging and ensuring convergence. The relaxed problem will be decomposed into two manageable subproblems, and these subproblems will then be coordinated by Lagrangian multipliers. 

\textbf{Step 1: Solve ``Loss Function" Subproblem for ${\bf{W}}_{n}$ by using Stochastic Gradient Descent.} At iteration $k$, for given values of multipliers ${\bf{\Lambda}}_{n}^k$, the first ``loss function" subproblem is to minimize the Lagrangian function while keeping ${\bf{Z}}_{n}$ at previously obtained values ${\bf{Z}}_{n}^{k-1}$ as
\begin{equation}
 \min_{ {\bf{W}}_{n}} L_{\rho} \big( {\bf{W}}_{n}, {\bf{Z}}_{n}^{k-1} , {\bf{\Lambda}}^k_{n} \big)
\label{lossfunctionsubproblem}
\end{equation}

Since the regularizer is a differentiable quadratic function, and the loss function is differentiable, the subproblem can be solved by stochastic gradient descent (SGD)~\cite{bottou2010large}.  To ensure that multiplier-updating directions are ``proper," the following ``surrogate" optimality condition needs to be satisfied following \cite[p. 179, eq. (12)]{Bragin2015SLR}:
\begin{equation}
\label{SOC1}
L_{\rho} \big({\bf{W}}_{n}^{k}, {\bf{Z}}_{n}^{k-1}, {\bf{\Lambda}}_{n}^{k} \big) < 
L_{\rho} \big({\bf{W}}_{n}^{k-1}, {\bf{Z}}_{n}^{k-1}, {\bf{\Lambda}}_{n}^{k} \big)
\end{equation}

If \eqref{SOC1} is satisfied, multipliers are updated following \cite[p. 179, eq. (15)]{Bragin2015SLR} as: 
\begin{equation} 
{\bf{\Lambda}'}_{n}^{k+1}: = {\bf{\Lambda}}_{n}^{k}+ s'^k({\bf{W}}_{n}^{k}-{\bf{Z}}_{n}^{k-1})
\label{multiplierupdate1}
\end{equation}
where stepsizes are updated as \cite[p. 180, eq. (20)]{Bragin2015SLR} as
\begin{equation}
s'^{k} = \alpha^{k} \frac{s^{k-1}||{\bf{W}}^{k-1}-{\bf{Z}}^{k-1}||}{||{\bf{W}}^{k}-{\bf{Z}}^{k-1}||}
\label{step1a}
\end{equation}

\textbf{Step 2: Solve ``Cardinality" Subproblem for ${\bf{Z}}_{n}$ through Pruning by using Projections onto Discrete Subspace.} The second ``cardinality" subproblem is solved with respect to ${\bf{Z}}_{n}$ while fixing other variables at values ${\bf{W}}_{n}^k$ as  
\begin{equation}
 \min_{ {\bf{Z}}_{n}} L_{\rho} \big( {\bf{W}}_{n}^k, {\bf{Z}}_{n}, {\bf{\Lambda}'}_{n}^{k+1} \big)
\label{cardinalitysubproblem}
\end{equation}

Since $g_{n}(\cdot)$ is an indicator function, the globally optimal solution of this problem can be explicitly derived as \cite{Boyd2011DistributedOA}:
\begin{equation}
\label{5}
  {\bf{Z}}_{n}^{k} = {{\bf{\Pi}}_{{\bf{S}}_{n}}}\big({\bf{W}}_{n}^{k}+\frac{{\bf{\Lambda}'}_{n}^{k+1}}{\rho}\big)
\end{equation}
where ${{\bf{\Pi}}_{{\bf{S}}_{n}}(\cdot)}$ denotes the Euclidean projection onto the set ${\bf{S}}_{n}= \{{\bf{W}}_{n}\mid \mathrm{card}({\bf{W}}_{n})\le l_{n}  \}, n=1,\dots,N$. To ensure that multiplier-updating directions are ``proper", the following ``surrogate" optimality condition needs to be satisfied:
\begin{equation}
\label{SOC}
 L_{\rho} \big({\bf{W}}_{n}^{k} , {\bf{Z}}_{n}^{k}, {\bf{\Lambda}'}_{n}^{k+1}  \big) < 
 L_{\rho} \big({\bf{W}}_{n}^{k} , {\bf{Z}}_{n}^{k-1} , {\bf{\Lambda}'}_{n}^{k+1} \big)
\end{equation}

Once \eqref{SOC} is satisfied,\footnote{If condition \eqref{SOC} is not satisfied, the subproblems \eqref{lossfunctionsubproblem} and \eqref{cardinalitysubproblem} are solved again by using the latest available values for ${\bf{W}}_{n}$ and ${\bf{Z}}_{n}$.} multipliers are updated as:
\begin{equation} 
{\bf{\Lambda}}_{n}^{k+1}: = {\bf{\Lambda}'}_{n}^{k+1}+ s^k({\bf{W}}_{n}^{k}-{\bf{Z}}_{n}^{k})
\label{multiplierupdate2}
\end{equation}
where stepsizes and stepsize-setting parameters \cite[p. 188, eq. (67)]{Bragin2015SLR} are updated as:
\begin{equation}
s^{k} = \alpha^{k} \frac{s'^{k}||{\bf{W}}^{k-1}-{\bf{Z}}^{k-1}||}{||{\bf{W}}^{k}-{\bf{Z}}^{k}||}; \quad \alpha^{k} = 1 - \frac{1}{M \times k^{(1-\frac{1}{k^r})}}, \: M > 1, 0 < r < 1
\label{step1b}
\end{equation}

The theoretical results are based on \cite{Gurevin2021EnablingRD, bragin2022novel, bragin2022surrogate} and are summarized below:

\textbf{Theorem 1 (Sufficient Condition for Convergence)} Assuming for any integer number $\kappa$ there exists $k > \kappa$ such that surrogate optimality conditions \eqref{SOC1} and \eqref{SOC} are satisfied, then  under the stepsizing conditions \eqref{step1a} and \eqref{step1b}, the Lagrangian multipliers converge to their optimal values $\mathbf{\Lambda}_n^{*}$ that maximize the following dual function:  
\begin{equation}
q({\bf{\Lambda}}) \equiv \min_{{\bf{W}}_{n} , {\bf{Z}}_{n}} L_{\rho} \big({\bf{W}}_{n} , {\bf{Z}}_{n}, {\mathbf{\Lambda}}_{n}  \big)
\label{maximizationofdualfunction}
\end{equation}

\begin{proof}
The proof will be based on that of \cite{Bragin2015SLR}. The major difference between the original SLR method \cite{Bragin2015SLR} and the SLR method of this paper is the presence of quadratic terms within the Lagrangian function \eqref{relaxedproblem}.  

It is important to mention that the weight pruning problem can be equivalently rewritten in a generic form as \eqref{GeneralEquation}, where $\bf{X}$ collectively denotes the decision variables $\{{\bf{W}}_{n}, {\bf{Z}}_{n}\}$ 
\begin{equation}
\min_{\bf{X}} \bf{F}(\bf{X}), 
\quad \textit{s.t.} \quad  \bf{G}(\bf{X}) = 0
\label{GeneralEquation}
\end{equation}
where 
\begin{equation}
\label{fff}
{\bf{F(X)}} \equiv f \big( {\bf{W}}_{n} \big)  + \sum_{n=1}^{N} g_{n}({\bf{Z}}_{n}) + \sum_{n=1}^{N} \frac{\rho}{2} \| {\bf{W}}_{n}-{\bf{Z}}_{n} \|_{F}^{2}, \quad {\bf{G(X)}} \equiv {\bf{W}}_{n} - {\bf{Z}}_{n}, \quad n = 1, \ldots, N
\end{equation}

The feasible set of \eqref{GeneralEquation} is equivalent to the original model compression problem. Feasibility requires ${\bf{W}}_{n} = {\bf{Z}}_{n}$, which makes the term $\frac{\rho}{2} \| {\bf{W}}_{n}-{\bf{Z}}_{n} \|_{F}^{2}$ within \eqref{fff} disappear. Therefore, the Lagrangian function corresponding to \eqref{GeneralEquation} is the \textit{Augmented} Lagrangian function \eqref{relaxedproblem} to the original model compression problem. Furthermore, the surrogate optimality conditions \eqref{SOC1} and \eqref{SOC} are the surrogate optimality conditions that correspond to the Lagrangian function $\bf{F(\bf{X})} + \bf{\Lambda} \bf{G(\bf{X})}$ that corresponds to \eqref{GeneralEquation}. Therefore, since within the original SLR \cite[Prop. 2.7, p.188]{Bragin2015SLR} convergence was proved under conditions on stepsizes \eqref{step1a} and \eqref{step1b} and the satisfaction of surrogate optimality conditions, which are assumed to be satisfied here, multipliers converge to their optimal values for the model compression under consideration as well.  
\end{proof}
\vspace{-0.8mm}
Although the convergence proof in Theorem 1 is valuable for distinguishing SLR from previous decomposition and coordination methods (e.g., ADMM) in terms of convergence, it does not provide insight into the solution quality on its own. 
This issue is addressed in Theorem 2, where the faster convergence of SLR is rigorously quantified. Since both ADMM and SLR methods are dual methods to maximize the dual function \eqref{maximizationofdualfunction}, it is common practice to determine upper bounds for the maximization problems. In the context of the problem being examined, an upper bound for the optimal dual value $q^*$ will be established within each method, enabling the evaluation of the quality of dual solutions - specifically, Lagrangian multipliers serve as the decision variables in the dual space.\footnote{Higher quality of primal variables - weights and biases is implied since superior coordination through Lagrangian multipliers significantly improves the quality of primal solutions. Though in a different problem context, this assertion has been empirically verified \cite{bragin2022surrogate}.}

\textbf{Theorem 2 (Dual solution quality: Best-case performance)}. Assuming that the ``sufficient condition for convergence'' stated within Theorem 1 is satisfied, then Surrogate Lagrangian Relaxation provides a better dual solution quality as compared to ADMM: in particular, there exists an iteration $\kappa$ so that for all iterations $k>\kappa$, the following condition holds:
\begin{equation} 
\overline{q}^{SLR}_{\kappa} < \overline{q}^{ADMM}_{\kappa}
\label{SLRbetterquality}
\end{equation}

\begin{proof}
There exists an overestimate of the optimal dual value \cite{bragin2022surrogate, bragin2022novel}, which in terms of our problem under consideration can be expressed as:
\vspace{-0.5mm}
\begin{equation} 
\overline{q}_{k}^{SLR} = \gamma \cdot s^k \cdot \| {\bf{W}}_{n}^k-{\bf{Z}}_{n}^k \|_{F}^{2} + L_{\rho} \big({\bf{W}}_{n}^k , {\bf{Z}}_{n}^k, {\mathbf{\Lambda}}_{n}^k  \big)
\label{eq27}
\end{equation}
Here $\gamma \in [0,1]$ is a parameter. Since within the surrogate Lagrangian relaxation, stepsizes are approaching 0, then
\begin{equation} 
\overline{q}_{k}^{SLR} \rightarrow  L_{\rho} \big({\bf{W}}_{n}^k , {\bf{Z}}_{n}^k, {\mathbf{\Lambda}}_{n}^k  \big)
\label{eq28}
\end{equation}
Moreover, assuming that a sufficient condition for convergence is satisfied, then the Lagrangian dual value approaches the dual value (as proved, for example, it \cite{Bragin2015SLR}), therefore, 
\begin{equation} 
\overline{q}_{k}^{SLR} \rightarrow  q(\lambda^*)
\label{eq29}
\end{equation}
Within ADMM, however, since stepsizes/penalty coefficients do not approach zero, then the overestimate of the optimal dual value $q(\lambda^*)$ is bounded away from it:
\begin{equation} 
\overline{q}_{k}^{ADMM} = \gamma \cdot \rho \cdot \| {\bf{W}}_{n}^k-{\bf{Z}}_{n}^k \|_{F}^{2} + L_{\rho} \big({\bf{W}}_{n}^k , {\bf{Z}}_{n}^k, {\mathbf{\Lambda}}_{n}^k  \big)
\label{eq30}
\end{equation}
Therefore, since the first term does not approach zero, and the second term does not approach the optimal dual value from above, there exists $\kappa$ such that \eqref{SLRbetterquality} holds.
\end{proof}
\begin{wrapfigure}{L}{0.5\textwidth}
\begin{minipage}{0.5\textwidth}
\begin{algorithm}[H] 
\caption{Surrogate Lagrangian Relaxation}
\label{algorithm}
\begin{algorithmic}[1] 
\STATE Initialize ${\bf{W}}_{n}^{0} , {\bf{Z}}_{n}^{0} , {\bf{\Lambda}}_{n}^{0} $ and $s^0$ \\
\WHILE{Stopping criteria are not satisfied} 
    \STATE \textbf 1 solve subproblem \eqref{lossfunctionsubproblem}, \;
    \IF{surrogate optimality condition \eqref{SOC1} is satisfied }
        \STATE keep ${\bf{W}}_{n}^{k} , {\bf{Z}}_{n}^{k} $, and update multipliers ${\bf{\Lambda}}_{n}^{k} $ per \eqref{multiplierupdate1},
    \ELSE
        \STATE  keep ${\bf{W}}_{n}^{k} , {\bf{Z}}_{n}^{k} $, do not update multipliers ${\bf{\Lambda}}_{n}^{k} $, 
    \ENDIF

    \STATE \textbf 2 solve subproblem \eqref{cardinalitysubproblem}, \; 
    \IF{surrogate optimality condition \eqref{SOC} is satisfied }
        \STATE keep ${\bf{W}}_{n}^{k} ,  {\bf{Z}}_{n}^{k} $, and update multipliers ${\bf{\Lambda}}_{n}^{k} $ per \eqref{multiplierupdate2},
    \ELSE
        \STATE  keep ${\bf{W}}_{n}^{k} ,  {\bf{Z}}_{n}^{k} $, do not update multipliers ${\bf{\Lambda}}_{n}^{k} $, 
   \ENDIF
\ENDWHILE
\end{algorithmic}
\end{algorithm}
\end{minipage}
\end{wrapfigure}

The algorithm of the proposed SLR method is summarized in Algorithm~\ref{algorithm}. 
The SLR method benefits from efficient subproblem solution coordination with guaranteed convergence enabled by stepsizes \eqref{step1a} and \eqref{step1b} approaching zero. Without this requirement, multipliers \eqref{multiplierupdate2} would not exhibit convergence. By the satisfaction of surrogate optimality conditions \eqref{SOC1} and \eqref{SOC} ensuring that multipliers are updated along ``good" directions. Section \ref{evaluation} will empirically verify that there always exists iteration $\kappa$ after which the ``surrogate'' optimality conditions are satisfied ensuring that multipliers approach their optimal values during the entire iterative process.

The SLR method also benefits from the independent and systematic adjustment of two hyper-parameters: penalty coefficient $\rho$ and the stepsize $s^k$. In contrast, other coordination methods are not designed to handle discrete variables and other types of non-convexities.  For example, ADMM does not converge when solving non-convex problems \cite[p.~73]{Boyd2011DistributedOA} because stepsizes $\rho$ within the method do not converge to zero. Lowering stepsizes to zero within ADMM would also result in a decreased penalty coefficient, leading to slower convergence.





\section{Evaluation} 
\label{evaluation}

In this section, we discuss our experimental results for the image classification task using CNN-based models and non-CNN-based models. We also evaluate our method under object detection and image segmentation tasks. 

\subsection{Experimental Setup}

All of our code, including image classification tasks and object detection and segmentation tasks, is implemented with Python 3.6 and PyTorch 1.6.0. For our experiments on the COCO 2014 dataset, we used the pycocotools v2.0 packages. For our experiments on TuSimple lane detection benchmark dataset\footnote{\url{https://github.com/TuSimple/tusimple-benchmark}}, we used SpConv v1.2 package. We conducted our experiments on Ubuntu 18.04 using Nvidia Quadro RTX 6000 GPU with 24 GB GPU memory. 
 
We begin by pruning the pretrained models through SLR training. Afterward, we perform hard-pruning on the model, completing the compression phase. We report the overall compression rate (or the percentage of remaining weights) and prediction accuracy.

\subsection{Evaluation of Image Classification Tasks using CNN-based Models} 
\label{classification}

\paragraph{\textbf{Models and Datasets.}} We use ResNet-18, ResNet-50, ResNet-56, ResNet-110 \cite{he2016deep} and VGG-16 \cite{simonyan2014} on CIFAR-10. On ImageNet ILSVRC 2012 benchmark, we use ResNet-18, ResNet-50 \cite{he2016deep} and MobileNetV2 \cite{Sandler2018MobileNetV2IR}. We use the pretrained ResNet models on ImageNet from Torchvision's ``models" subpackage.

\begin{table*}[!hb]
\caption{Comparison of SLR and ADMM on CIFAR-10 and ImageNet. ImageNet results show Top-1 / Top-5 accuracy.}
\label{table:addmvsslr_classification}
\centering
\begin{tabular}{cccccc}
\hline\hline
 & \textbf{Baseline (\%)} & \textbf{Epoch} & \textbf{ADMM (\%)} & \textbf{SLR (\%)} & \textbf{Compression Rate} \\ 
\hline
\textbf{CIFAR-10} &    &   &    &  & \\
\hline
{ResNet-18} &   {93.33}  & \multicolumn{1}{c|}{40}  & 72.84  & \textit{89.93}& {8.71$\times$} \\ 
\hline                       
{ResNet-50} &   {93.86}  & \multicolumn{1}{c|}{50}  & 78.63  & \textit{88.91}& {6.57$\times$} \\  
\hline
{VGG-16} &   {93.27}  & \multicolumn{1}{c|}{110}  & 69.05  & \textit{87.31}& {12$\times$} \\
\hline
{ResNet-56} &   {93.39}  & \multicolumn{1}{c|}{30}  & 90.5  & \textit{92.3}& {6.5$\times$} \\  
\hline
{ResNet-110} &   {93.68}  & \multicolumn{1}{c|}{30}  & 89.71  & \textit{92.31}& {9.7$\times$} \\ 
\hline\hline                 
\textbf{ImageNet} &    &   &    &  & \\
\hline
{ResNet-18} &   {69.7 / 89.0}  & \multicolumn{1}{c|}{40}  & 58.9 / 81.7  & \textit{60.9 / 84.4 }& {6.5$\times$} \\
\hline
{ResNet-50} &   {76.1 / 92.8}  & \multicolumn{1}{c|}{30}  & 64.8 / 85.1    & \textit{65.9 / 87.5}& {3.89$\times$} \\  
\hline
{MobileNetV2} &   {71.8 / 91.0 }  & \multicolumn{1}{c|}{60}  & 61.8 / 84.3    & \textit{63.2 / 85.5}& {1.76$\times$} \\ 
\hline\hline                        
\end{tabular}
\end{table*}

\begin{figure}[!ht]
  \centering
  \subcaptionbox{\centering{ ResNet-18 on CIFAR-10}. \label{fig:resnet18_cifar}}{\includegraphics[width=0.3\columnwidth]{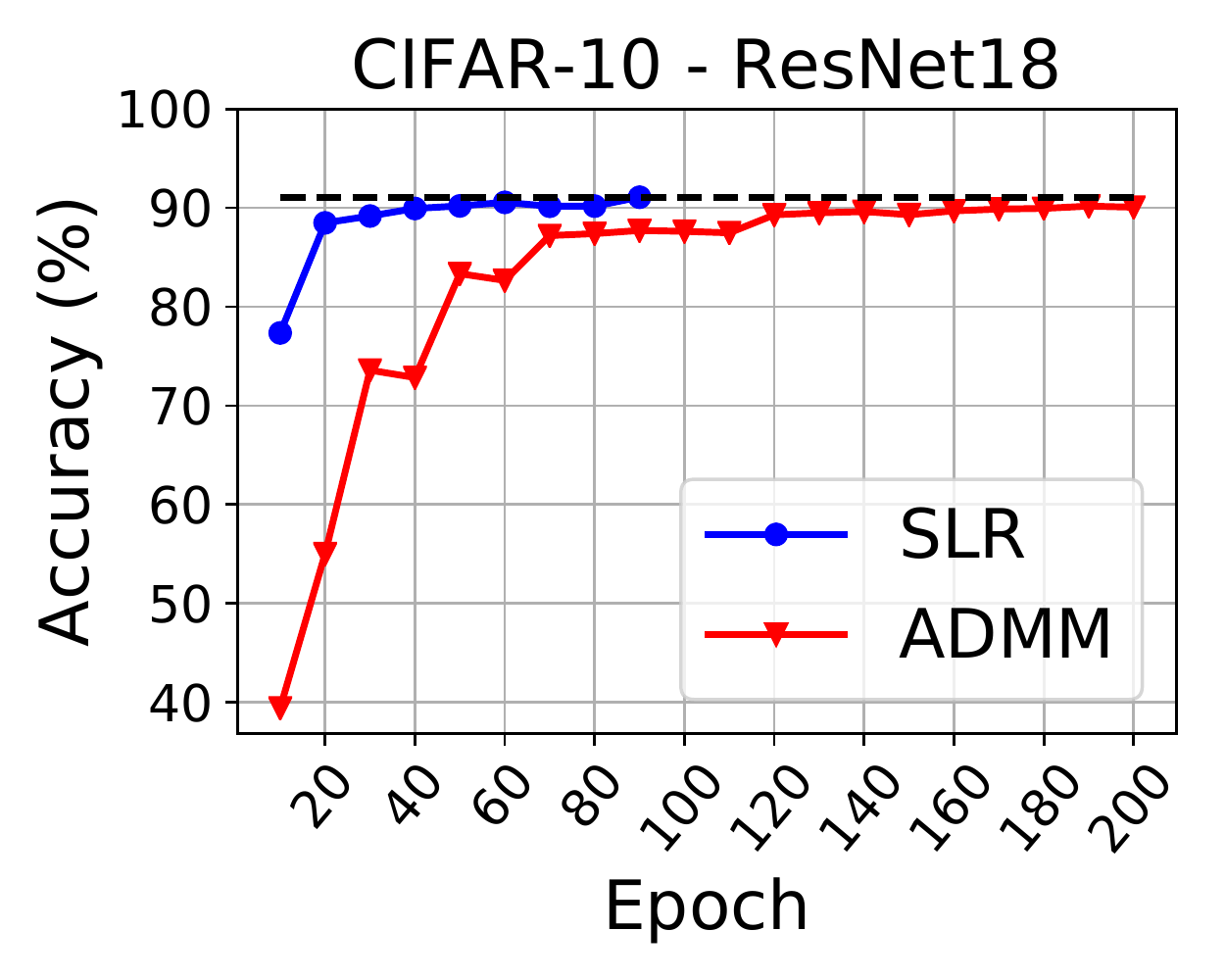}}\hspace{1em}
  \subcaptionbox{\centering{ ResNet-50 on CIFAR-10}. \label{fig:resnet50_cifar}}{\includegraphics[width=0.3\columnwidth]{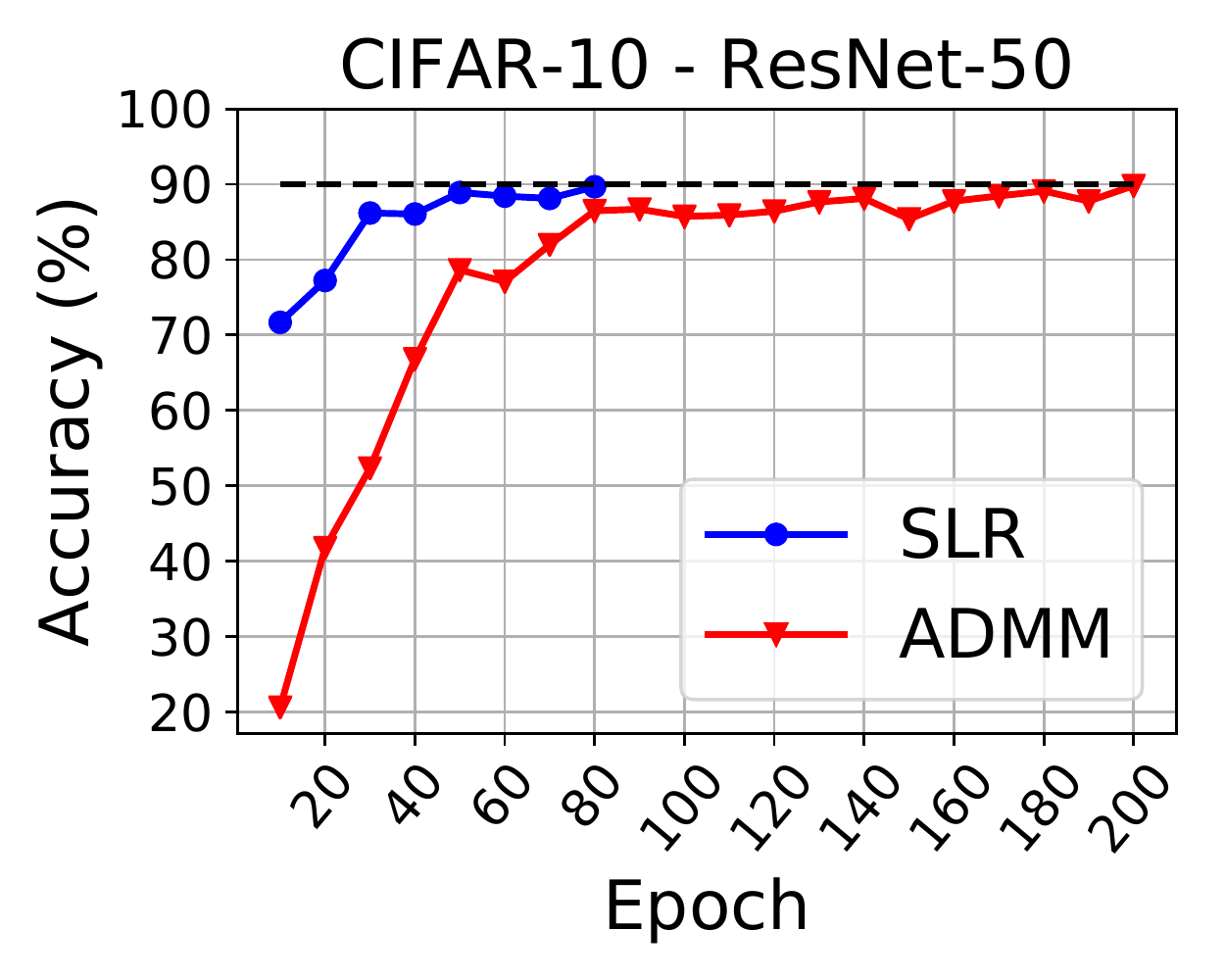}}\hspace{1em}
  \subcaptionbox{\centering{ VGG-16 on CIFAR-10}.\label{fig:vgg_cifar}}{\includegraphics[width=0.3\columnwidth]{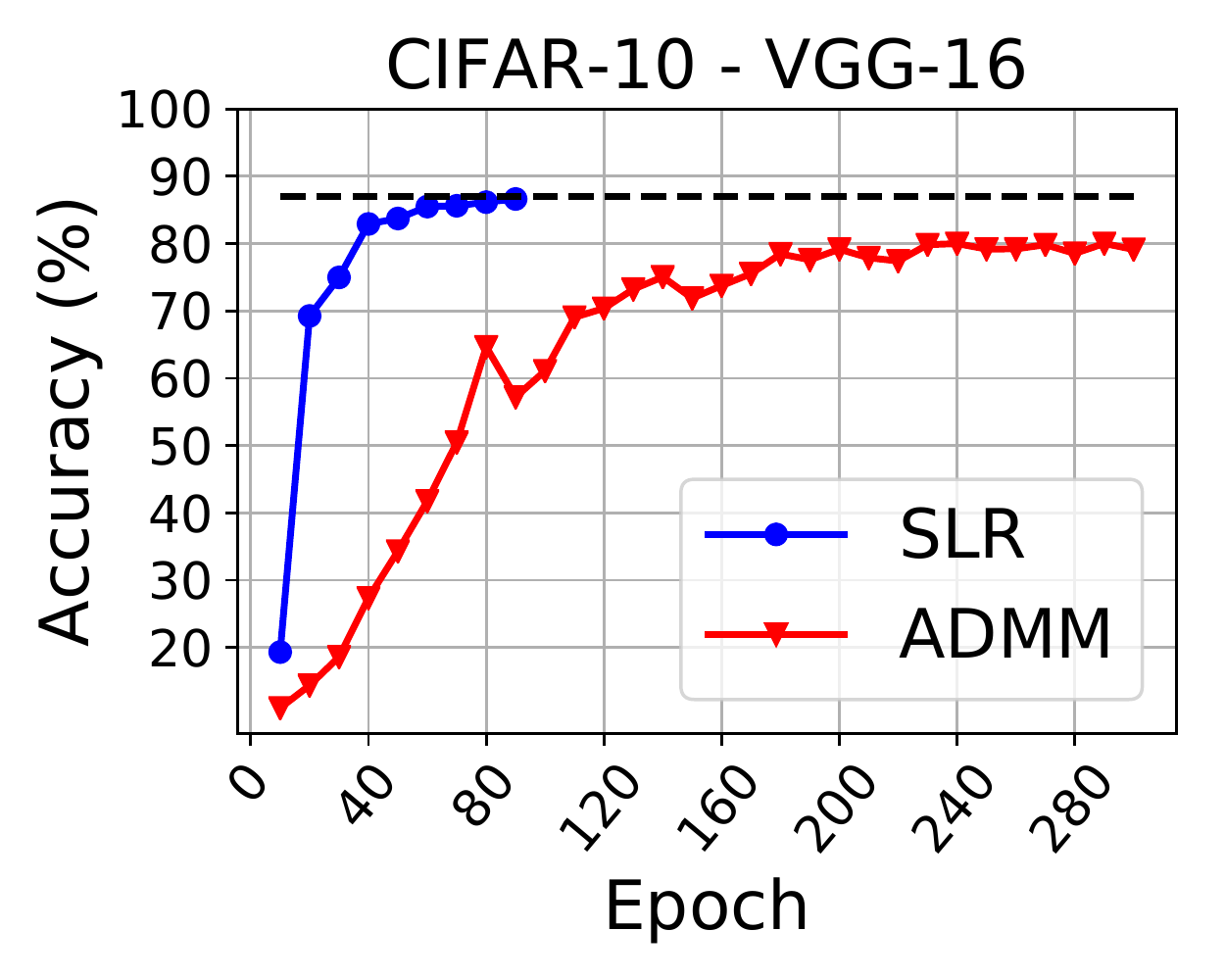}}
  \subcaptionbox{\centering{ ResNet-56 on CIFAR-10}. \label{fig:resnet56_cifar}}{\includegraphics[width=0.24\columnwidth]{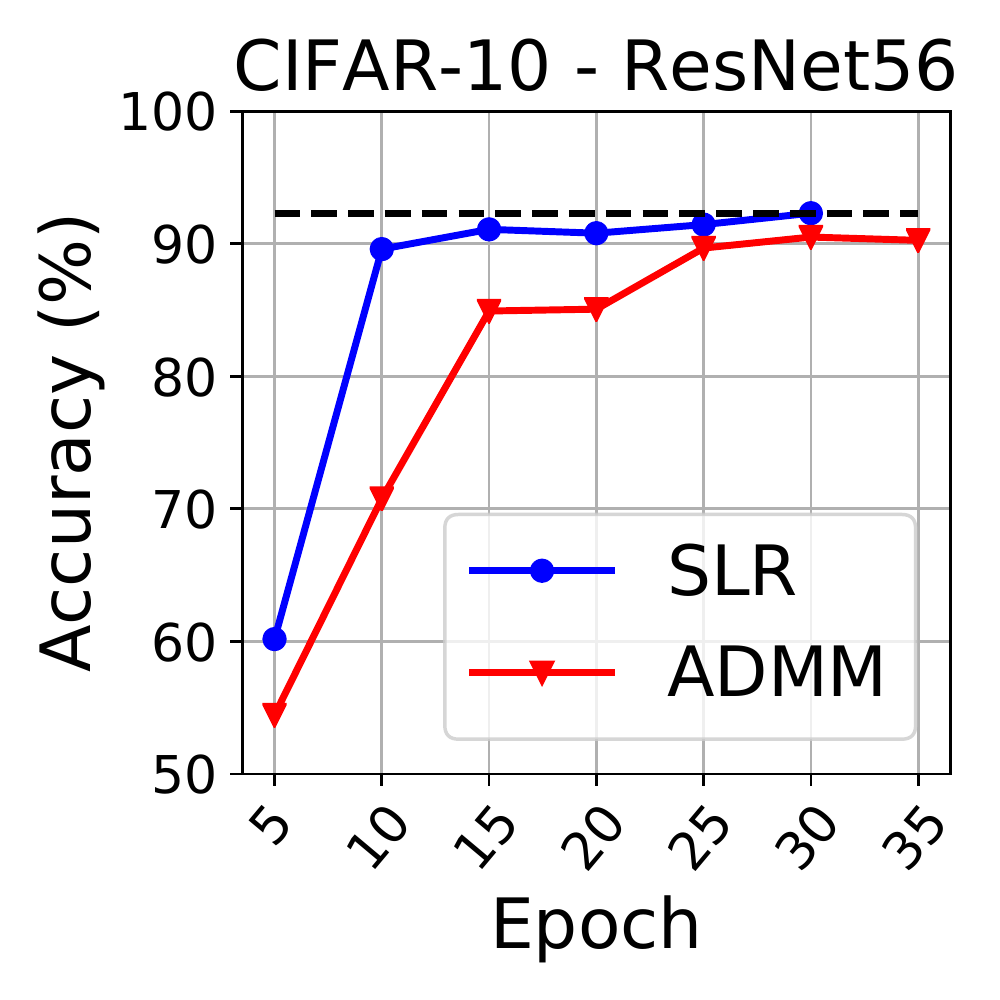}}\hspace{0em}
  \subcaptionbox{\centering{ResNet-110 on CIFAR-10}. \label{fig:resnet110_cifar}}{\includegraphics[width=0.24\columnwidth]{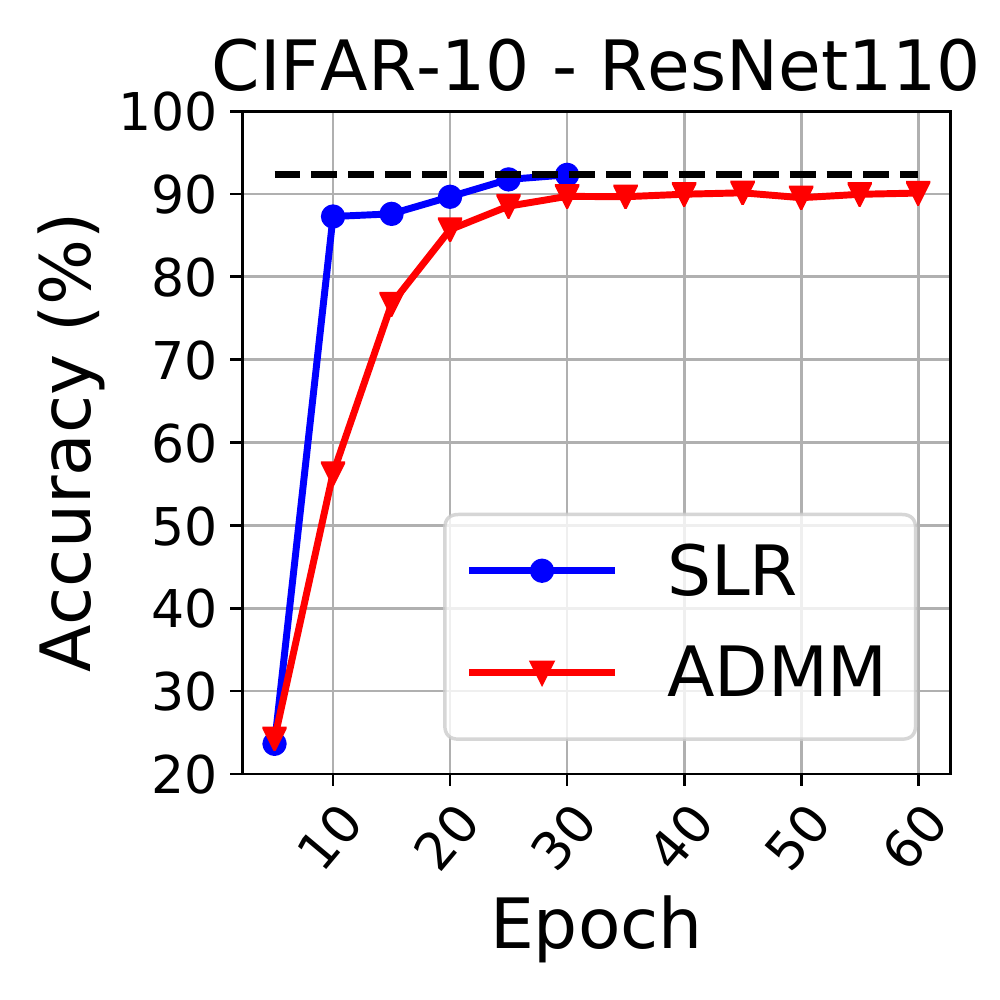}}\hspace{0em}%
  \subcaptionbox{\centering{ ResNet-18 on ImageNet}. \label{fig:resnet18_imagenet}}{\includegraphics[width=0.24\columnwidth]{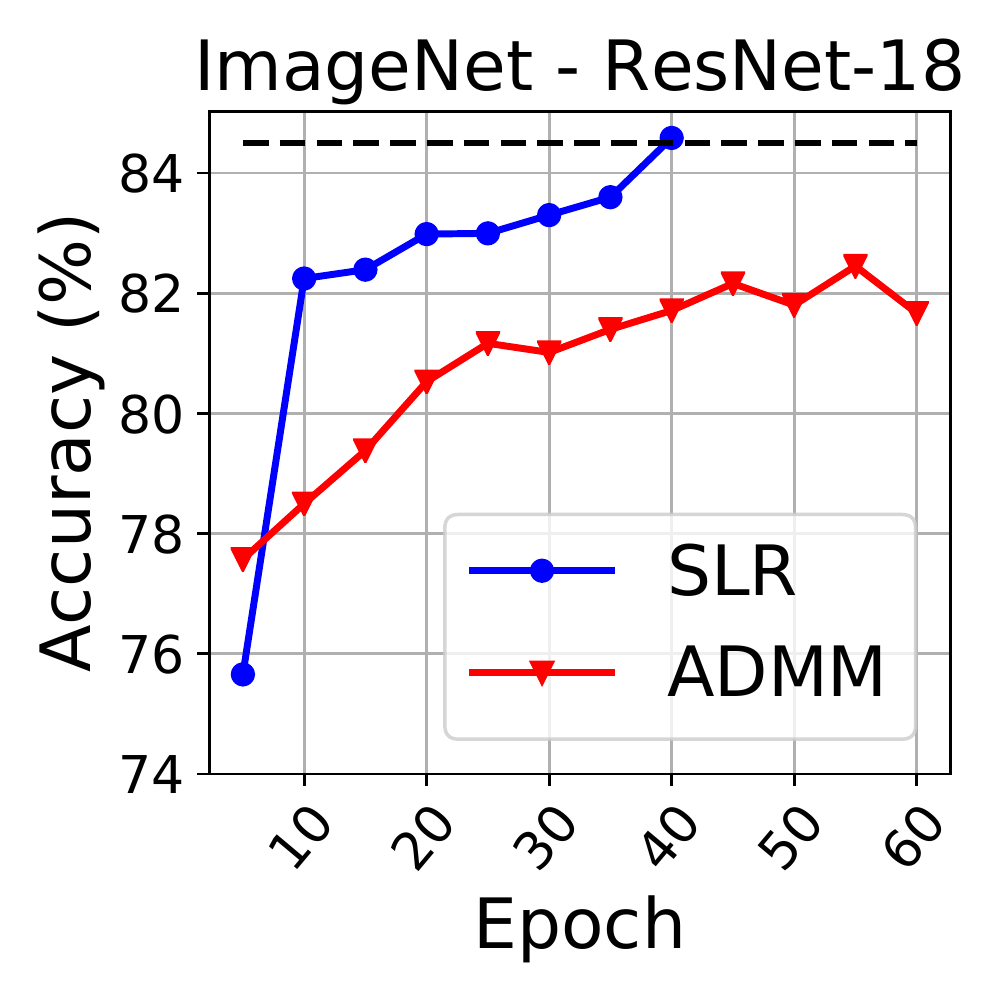}}\hspace{0em}%
  \subcaptionbox{\centering{ ResNet-50 on ImageNet}. \label{fig:resnet50_imagenet}}{\includegraphics[width=0.24\columnwidth]{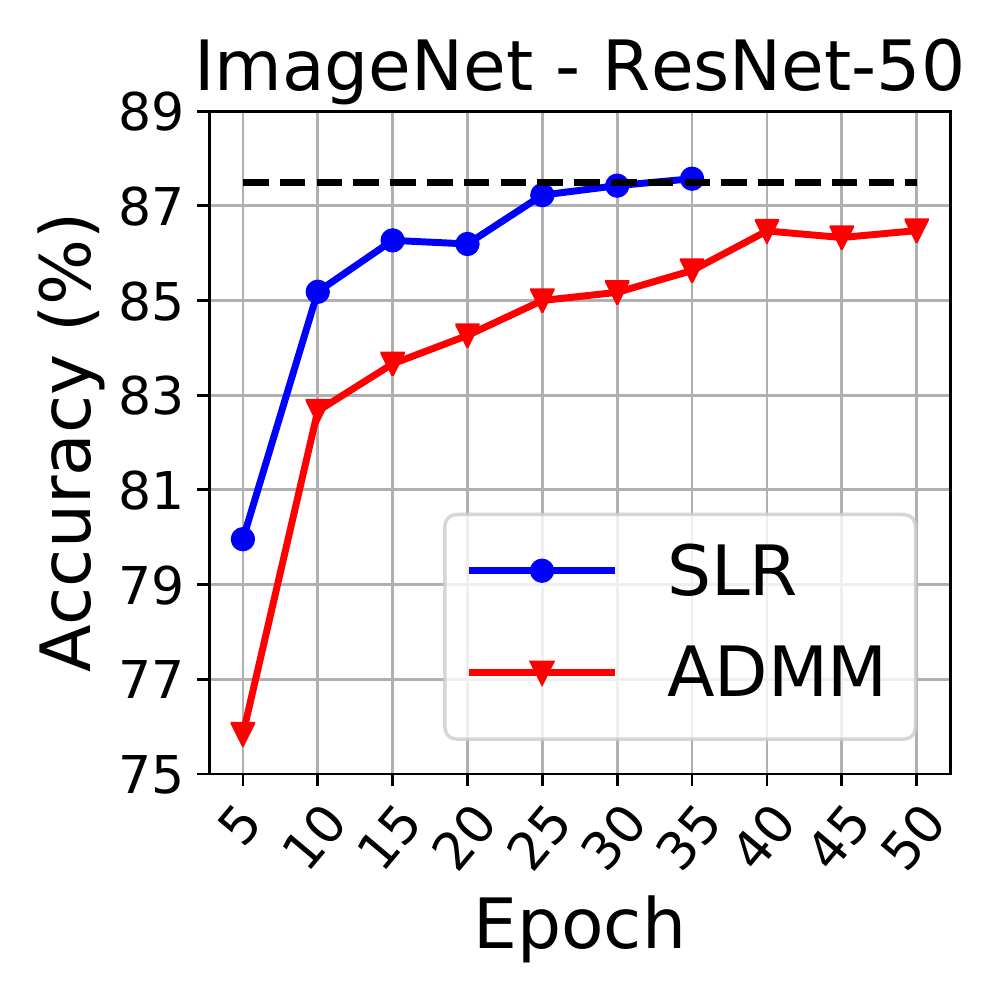}}
  \caption{Hardpruning accuracy after SLR and ADMM training on CIFAR-10 and ImageNet benchmarks. Accuracy is reported periodically and training is stopped when desired accuracy is reached. }
\label{fig:hardprune_cifar10}
\end{figure}

\paragraph{\textbf{Training Settings.}} In all the experiments, we set $\rho = 0.1$. In CIFAR-10 experiments, we use a learning rate of $0.01$, batch size of $128$, and the ADAM optimizer. In ImageNet experiments, we use a learning rate of $10^{-4}$, batch size of $256$, and the SGD optimizer. For a fair comparison of SLR and ADMM methods, we use the same number of training epochs and sparsity configurations for both methods in the experiments.

Table~\ref{table:addmvsslr_classification} shows our comparison of SLR and ADMM on CIFAR-10 and ImageNet benchmarks. For both experiments, SLR parameters are set as $M = 300$, $r = 0.1$, and $s_0 = 10^{-2}$. After the SLR and ADMM training, hardpruning is performed, and the hardpruning accuracy is reported without any additional retraining, given a limited budget of training epochs. According to our results, SLR outperforms the ADMM method in terms of accuracy under the same compression rate. When the compression rate is the same, SLR always obtains higher classification accuracy compared with ADMM under the same number of epochs of training. As compression rates increase, the results reveal a more significant accuracy gap between SLR and ADMM across various network architectures. This demonstrates that our SLR converges faster and quickly recovers the model accuracy at the hardprune stage, which achieves retrain-free given the limited training budget.

Figure \ref{fig:hardprune_cifar10} shows the hardpruning accuracy for SLR vs. ADMM on CIFAR-10 and ImageNet, corresponding to Table \ref{table:addmvsslr_classification}. During training, hardpruning accuracy is checked periodically. If the hardpruning accuracy meets the accuracy criteria, the training is stopped. As seen in Figure \ref{fig:hardprune_cifar10}, SLR converges quickly and reaches the desired accuracy almost $3\times$ faster than ADMM on CIFAR-10. Moreover, in Figure \ref{fig:vgg_cifar}, ADMM is still below the desired accuracy even after $300$ epochs of training on VGG-16, while SLR finishes training at $80$ epochs. Similarly, as shown in Figures \ref{fig:resnet18_imagenet} and \ref{fig:resnet50_imagenet}, ADMM cannot achieve the desired accuracy after $60$ and $50$ epochs of training on ImageNet, while SLR reaches the threshold quicker.

\begin{table*}[!ht]
\caption{SLR performance comparison with VGG-16, ResNet-18, ResNet-50 and ResNet-56 on CIFAR-10.}
\label{table:slr_comp}
\centering
\begin{tabular}{llcc}
\hline\hline
\textbf{Model} & \textbf{Method} & \textbf{Accuracy (\%)} & \textbf{Params Pruned (\%)} \\ 
\hline
\multirow{7}{*}{VGG-16}    & \textbf{SLR} & 91.2 & \multirow{3}{*}{90} \\
                           & AMC \cite{He_2018_ECCV} & 91.0 &          \\
                           & L0 \cite{louizos2018learning} & 80.0 &    \\ \cline{2-4} 
                           & \textbf{SLR} & 93.1 & \multirow{2}{*}{60} \\
                           & One-shot pruning\cite{liu2018rethinking} & 92.4 & \\ 
                           \cline{2-4} 
                           & \textbf{SLR} & 93.2 & \multirow{2}{*}{50} \\
                           & Iter. Prun. \cite{han2015learning} & 92.2 & \\
\hline
\multirow{2}{*}{ResNet-18} & \textbf{SLR} \textit{(at 20k iterations)} & 89.9 & \multirow{2}{*}{88.6}                                              \\
                           & Iter. prun. \cite{frankle2018lottery} & 75.0 & \\
\hline
\multirow{2}{*}{ResNet-50} & \textbf{SLR} & 93.6 & \multirow{2}{*}{60} \\
                           & AMC \cite{He_2018_ECCV} & 93.5 & \\ 
\hline
\multirow{9}{*}{ResNet-56} & \textbf{SLR} & 92.3 & 84.4 \\
                           & GSM \cite{Ding2019GlobalSM} & 94.1 & 85.0 \\
                           & Group Sparsity \cite{Li2020GroupST} & 92.65 & 79.2 \\
                           &  \cite{Zhao2019VariationalCN} & 92.26 & 20.49 \\
                           & GAL-0.6 \cite{Lin2019TowardsOS} & 93.38 & 11.8 \\
                           & \cite{li2016pruning} & 93.06 & 13.7 \\
                           & NISP \cite{Yu2018NISPPN} & 93.01 & 42.6 \\
                           & KSE \cite{Li2019ExploitingKS} & 93.23 & 54.73 \\
                           & DHP-50 \cite{Li2020DHPDM} & 93.58 & 41.58 \\ 
\hline\hline
\end{tabular}
\end{table*}

Table \ref{table:slr_comp} shows the comparison of SLR with other recent model compression works on the CIFAR-10 benchmark. We report the percentage of parameters pruned after SLR training and the final accuracy. We start training the networks with a learning rate of $0.1$ and decrease the learning rate by a factor of $10$ at epochs $80$ and $100$. On ResNet-18, we compare our results at only 20k iterations. For VGG-16 and ResNet-50, we observe that SLR can achieve up to $60\%$ pruning with less than $1\%$ accuracy drop.


\subsection{Evaluation on Image Classification Tasks using state-of-the-art Non-CNN-based Models}
\label{NLP_classification}

In this subsection, we demonstrate our experimental results of applying the SLR weight pruning method on a multi-layer perceptron (MLPs)-based architecture: MLP-Mixer~\cite{tolstikhin2021mlp} and on an attention-based network: Swin Transformer~\cite{liu2021swin}.

\paragraph{\textbf{Models and Datasets.}} MLP-Mixer does not contain any convolution layer or self-attention block. This architecture relies solely on multi-layer perceptrons. Alternating between channel-mixing MLPs and token (image patch)-mixing MLPs, the MLP-Mixer can achieve decent accuracy with much fewer computational resources than state-of-the-art methods. Swin Transformer is a hierarchical transformer-based architecture that utilizes shifted windows for representation. It has linear computational complexity with respect to input image size and achieves state-of-the-art performance on both image classification and object detection and semantic segmentation tasks.
We conduct experiments on ImageNet ILSVRC 2012 benchmark. We utilize the pretrained models from the PyTorch Image Models code base~\cite{rw2019timm}.

\paragraph{\textbf{Training Settings.}} Similar to experiments with CNN-based models, we set $\rho=0.1$. We set the learning rate at $0.01$, batch size of $128$, and use SGD optimizer with the momentum of $0.9$ and a weight-decay of $10^{-4}$. To ensure a fair comparison, we used the same number of training epochs and sparsity configurations for both SLR and ADMM methods.

\begin{table}[!b]
\caption{SLR pruning results with MLP-Mixer on ImageNet through different compression rates.}
\label{table:mlp}
\centering
\begin{tabular}{cc|c|cccc|cccc}
\hline\hline
\multicolumn{2}{c|}{\textbf{\begin{tabular}[c]{@{}c@{}}Baseline\end{tabular}}}
& \textbf{\begin{tabular}[c]{@{}c@{}}Compression\end{tabular}}
& \multicolumn{4}{c|}{\textbf{\begin{tabular}[c]{@{}c@{}}Hardpruning Acc. (\%)\end{tabular}}} 
& \multicolumn{4}{c}{\textbf{\begin{tabular}[c]{@{}c@{}}Retraining Acc. (\%)\end{tabular}}} \\ 
\cline{4-11} 
\multicolumn{2}{c|}{\textbf{\begin{tabular}[c]{@{}c@{}}Acc (\%)\end{tabular}}} 
& \textbf{\begin{tabular}[c]{@{}c@{}}Rate\end{tabular}}
& \multicolumn{2}{c}{\textbf{\begin{tabular}[c]{@{}c@{}}ADMM\end{tabular}}}  
& \multicolumn{2}{c|}{\textbf{\begin{tabular}[c]{@{}c@{}}SLR\end{tabular}}} 
& \multicolumn{2}{c}{\textbf{\begin{tabular}[c]{@{}c@{}}ADMM\end{tabular}}} 
& \multicolumn{2}{c}{\textbf{\begin{tabular}[c]{@{}c@{}}SLR\end{tabular}}} \\ 
\hline 
Top@1  & Top@5  &              & Top@1  & Top@5  & Top@1  & Top@5  & Top@1  & Top@5  & Top@1  & Top@5   \\ 
\hline 
       &        & $1.96\times$ & 75.332 & 91.734 & 75.392 & 91.752 & 75.332 & 91.714 & 75.392 & 91.752     \\ 
\cline{4-11} 
76.598 & 92.228 & $3.16\times$ & 71.798 & 89.64  & 72.578 & 90.136 & 73.362 & 90.672 & 73.56  & 90.750     \\
\cline{4-11} 
       &        & $8.28\times$ & 47.592 & 71.698 & 54.834 & 77.696 & 70.864 & 89.38  & 71.036 & 89.494     \\
\hline\hline
\end{tabular}
\end{table}
\begin{table}[!b]
\caption{SLR pruning results with Swin Transformer (Tiny) on ImageNet through different compression rates.}
\label{table:swin}
\centering
\begin{tabular}{cc|c|cccc|cccc}
\hline\hline
\multicolumn{2}{c|}{\textbf{\begin{tabular}[c]{@{}c@{}}Baseline\end{tabular}}}
& \textbf{\begin{tabular}[c]{@{}c@{}}Compression\end{tabular}}
& \multicolumn{4}{c|}{\textbf{\begin{tabular}[c]{@{}c@{}}Hardpruning Acc. (\%)\end{tabular}}} 
& \multicolumn{4}{c}{\textbf{\begin{tabular}[c]{@{}c@{}}Retraining Acc. (\%)\end{tabular}}} \\ 
\cline{4-11} 
\multicolumn{2}{c|}{\textbf{\begin{tabular}[c]{@{}c@{}}Acc (\%)\end{tabular}}} 
& \textbf{\begin{tabular}[c]{@{}c@{}}Rate\end{tabular}}
& \multicolumn{2}{c}{\textbf{\begin{tabular}[c]{@{}c@{}}ADMM\end{tabular}}}  
& \multicolumn{2}{c|}{\textbf{\begin{tabular}[c]{@{}c@{}}SLR\end{tabular}}} 
& \multicolumn{2}{c}{\textbf{\begin{tabular}[c]{@{}c@{}}ADMM\end{tabular}}} 
& \multicolumn{2}{c}{\textbf{\begin{tabular}[c]{@{}c@{}}SLR\end{tabular}}} \\ 
\hline 
Top@1  & Top@5  &              & Top@1  & Top@5  & Top@1  & Top@5  & Top@1  & Top@5  & Top@1  & Top@5   \\ 
\hline 
       &        & $1.95\times$ & 78.910 & 94.288 & 79.018 & 94.446 & 79.146 & 94.424 & 79.108 & 94.456     \\ 
\cline{4-11} 
81.350 & 95.532 & $3.13\times$ & 73.074 & 91.370 & 74.322 & 91.988 & 75.278 & 92.55 & 75.508 & 92.630    \\
\cline{4-11} 
       &        & $4.50\times$ & 63.892 & 85.104 & 67.398 & 87.494 & 72.626 & 90.966 & 73.046 & 91.364     \\
\hline\hline
\end{tabular}
\end{table}

\paragraph{\textbf{Comparison of SLR and ADMM}} The comparison of SLR and ADMM applied on MLP-Mixer model on ImageNet benchmark is presented in Table~\ref{table:mlp}. We compare the two methods using three distinct compression rates. The final hardpruning accuracy is reported after $100$ epochs of training without further retraining. Retrain accuracy is also reported after $50$ epochs.
For all SLR experiments, we set the parameters to $M = 300$, $r = 0.1$, and $s_0 = 0.01$. As indicated in Table~\ref{table:mlp}, in terms of hardpruning, when the compression rate is low as $1.96\times$, 
the two methods have similar performance.
As the compression rate increases to $3.16\times$ and $8.28\times$, SLR outperforms ADMM, with the accuracy gap widening as the compression rate increases. This demonstrates that SLR outperforms ADMM by achieving higher accuracy during the hardpruning stage, which leads to more efficient use of training resources.

The comparison of SLR and ADMM on the Swin Transformer model on ImageNet benchmark is shown in Table~\ref{table:swin}. In all the experiments, we choose the tiny version (Swin-T) as it balances the model size and accuracy very well. For SLR experiments, when compression rate is $1.96\times$, we use parameters as $M = 150$, $r = 0.1$ and $s_0 = 0.05$. When compression rate is $3.16\times$, we use parameters as $M = 300$, $r = 0.1$ and $s_0 = 0.01$. And when compression rate is $4.5\times$, we use parameters as $M = 150$, $r = 0.05$ and $s_0 = 0.005$. As demonstrated, when the compression rate is $1.95\times$, SLR obtains higher accuracy than ADMM at the hardpruning stage. After $50$ epochs of retraining, the accuracy of ADMM improves $0.2\%$ and becomes closer to the accuracy of SLR, while SLR improves $0.01\%$. This demonstrates that our SLR can obtain higher accuracy at the hardpruning stage. As the compression rate increases, the accuracy gap between SLR and ADMM also widens. For instance, at a compression rate of $4.5\times$, the accuracy of SLR is $2.4\%$ higher than that of ADMM at the hardpruning stage.

\begin{figure*}[!b]
  \centering
   \subcaptionbox{Compression rate = $1.96\times$. \label{fig:mlp_0.5}}{\includegraphics[width=0.32\columnwidth]{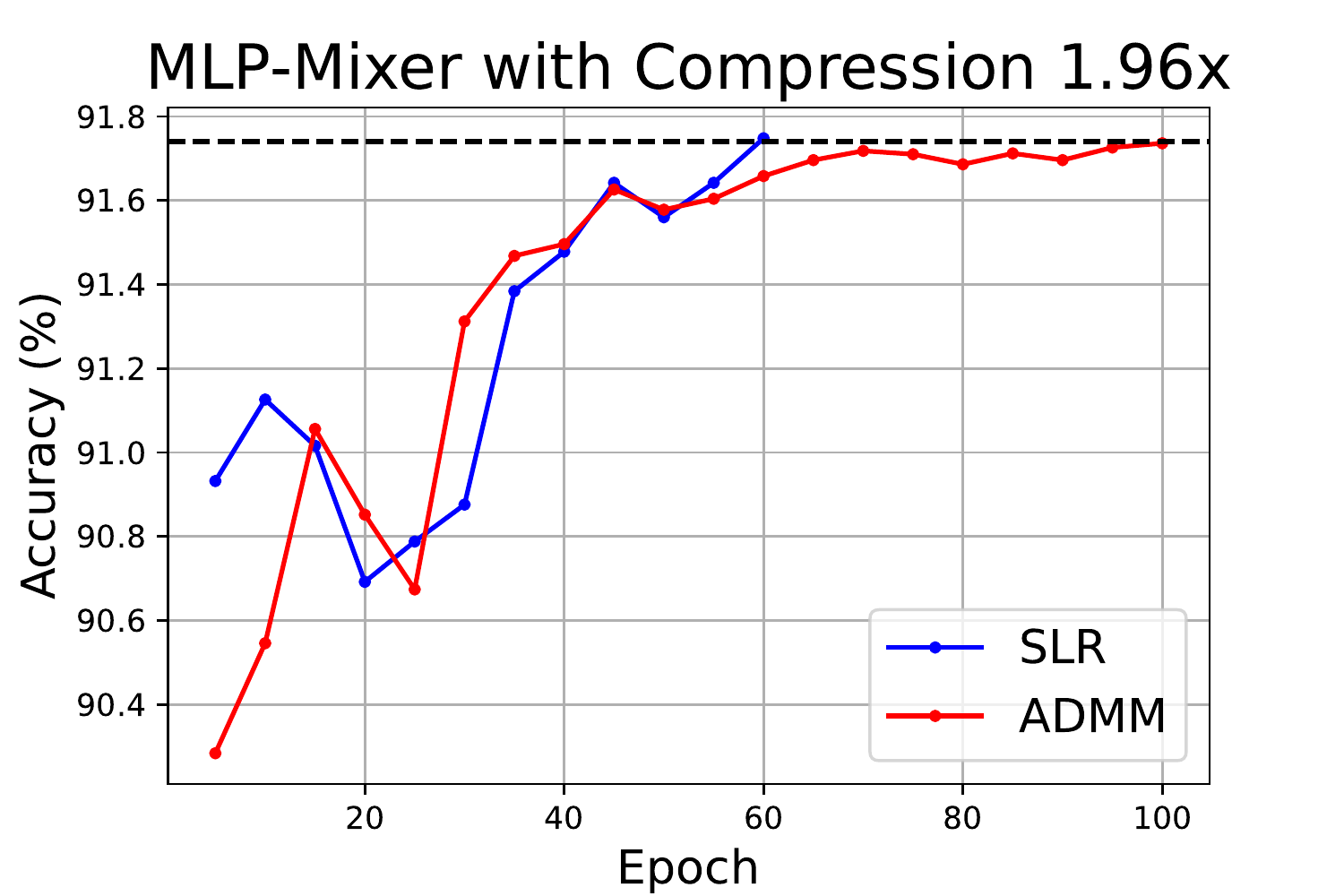}}
  \subcaptionbox{Compression rate = $3.16\times$.
  \label{fig:mlp_0.7}}{\includegraphics[width=0.32\columnwidth]{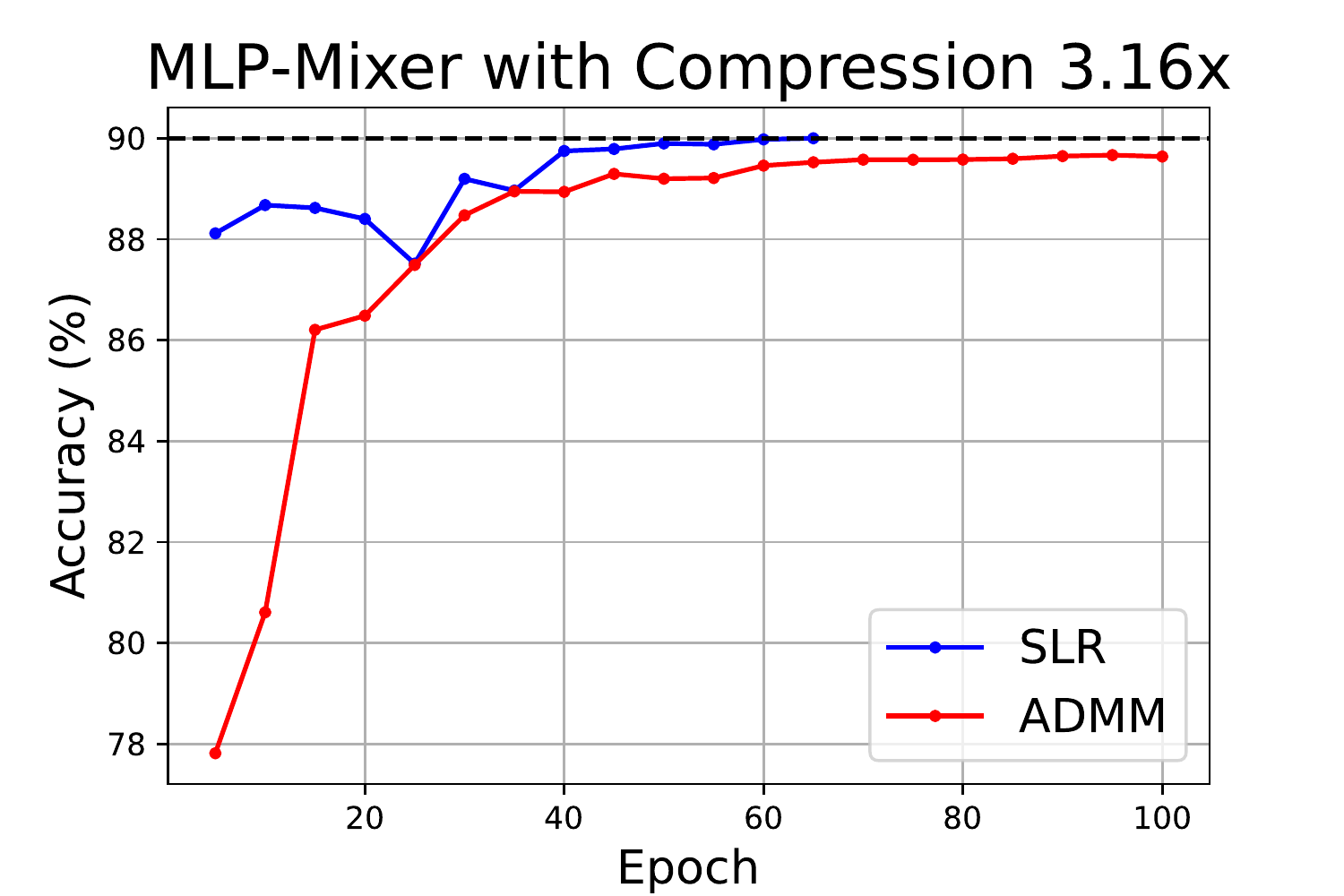}}\hspace{1em}%
  \subcaptionbox{Compression rate = $8.28\times$. \label{fig:mlp_0.9}}{\includegraphics[width=0.32\columnwidth]{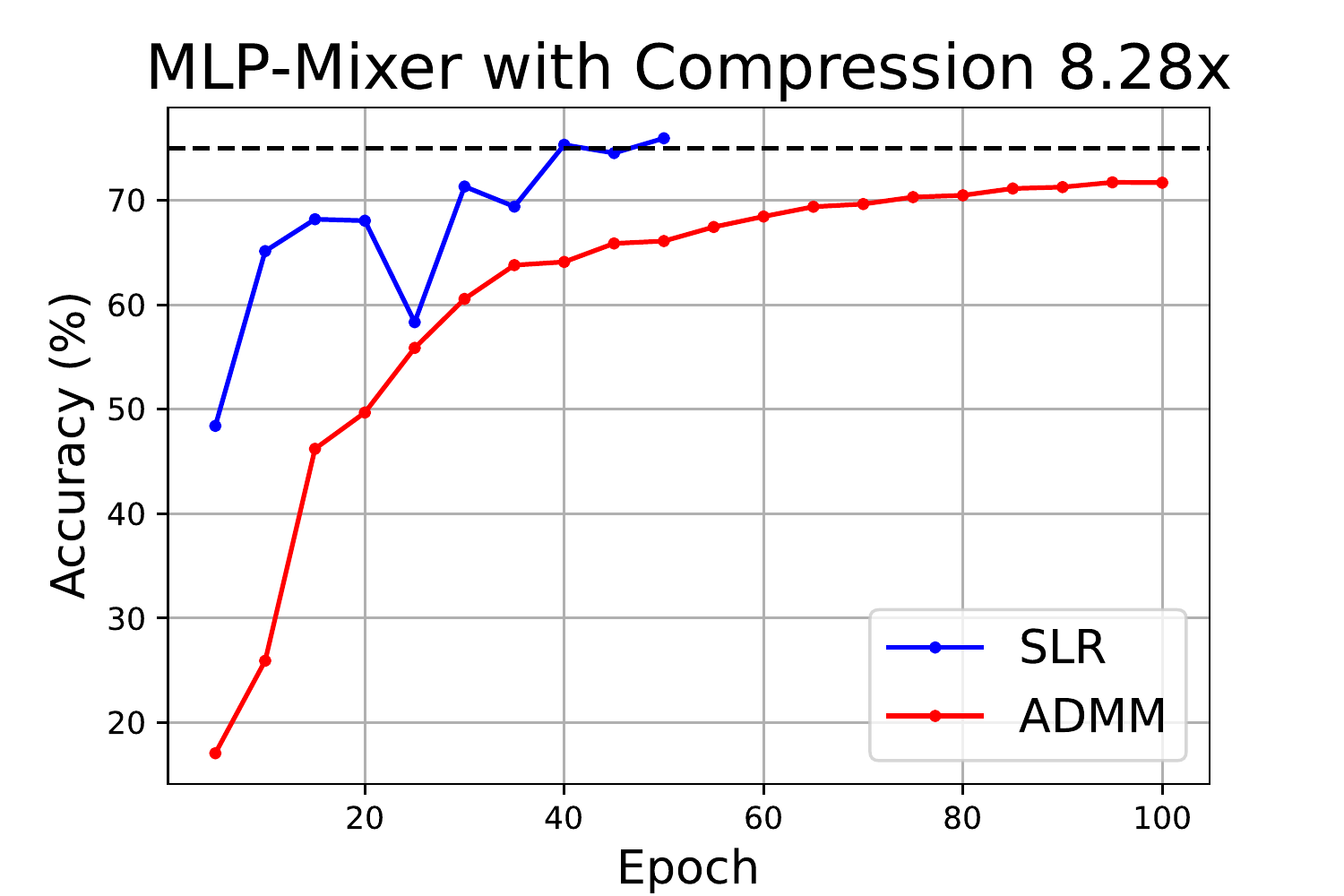}}
  \caption{Top5 Hardprune accuracy of MLP-Mixer on ImageNet after SLR and ADMM pruning. Accuracy is reported periodically.}
\label{fig:hardprune_mlp}
\end{figure*}
\begin{figure*}[!b]
  \centering
   \subcaptionbox{Compression rate = $1.95\times$. \label{fig:swin_0.5}}{\includegraphics[width=0.3\columnwidth]{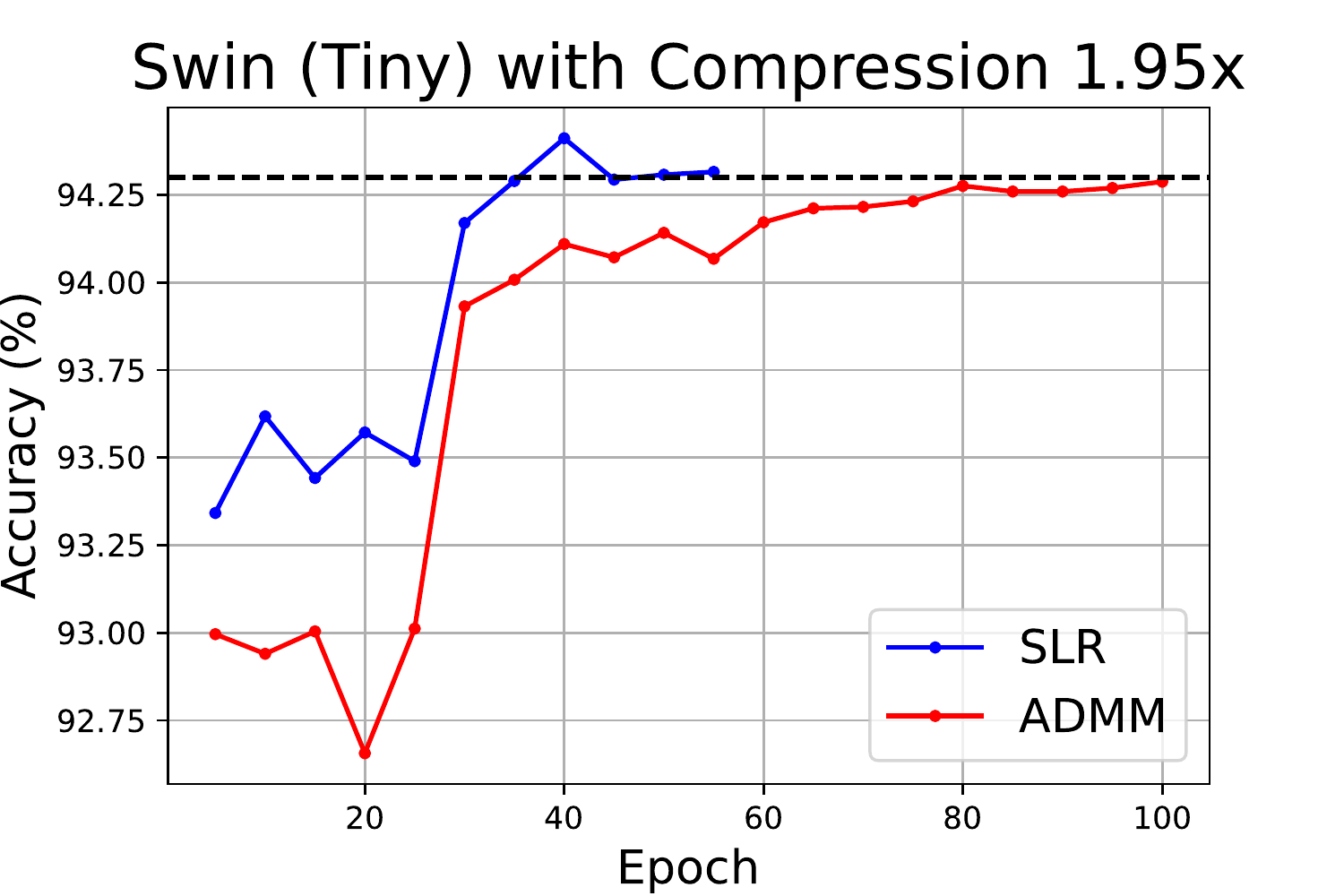}}
  \subcaptionbox{Compression rate = $3.13\times$.
  \label{fig:swin_0.7}}{\includegraphics[width=0.3\columnwidth]{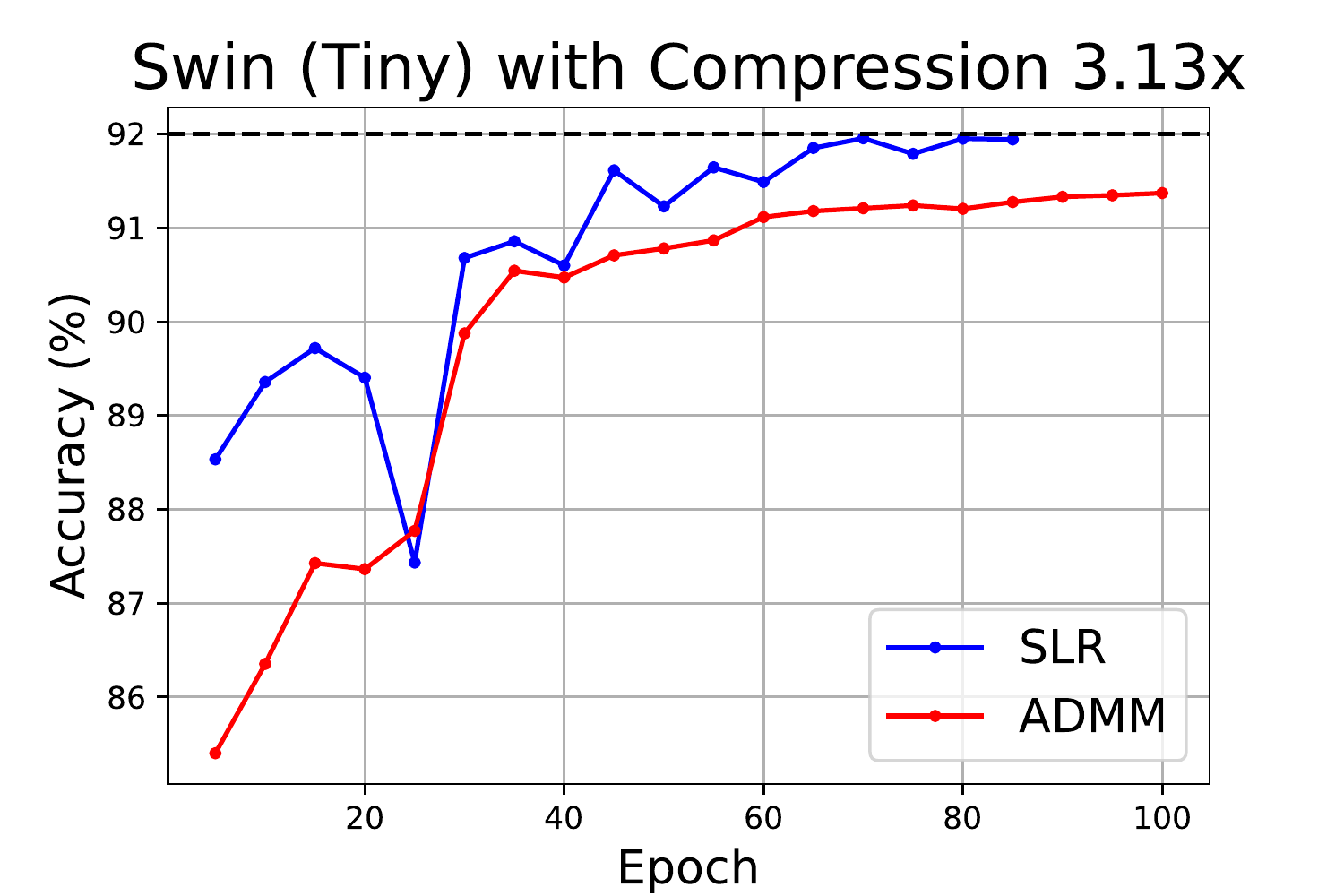}}\hspace{1em}%
  \subcaptionbox{Compression rate = $4.5\times$. \label{fig:swin_0.8}}{\includegraphics[width=0.3\columnwidth]{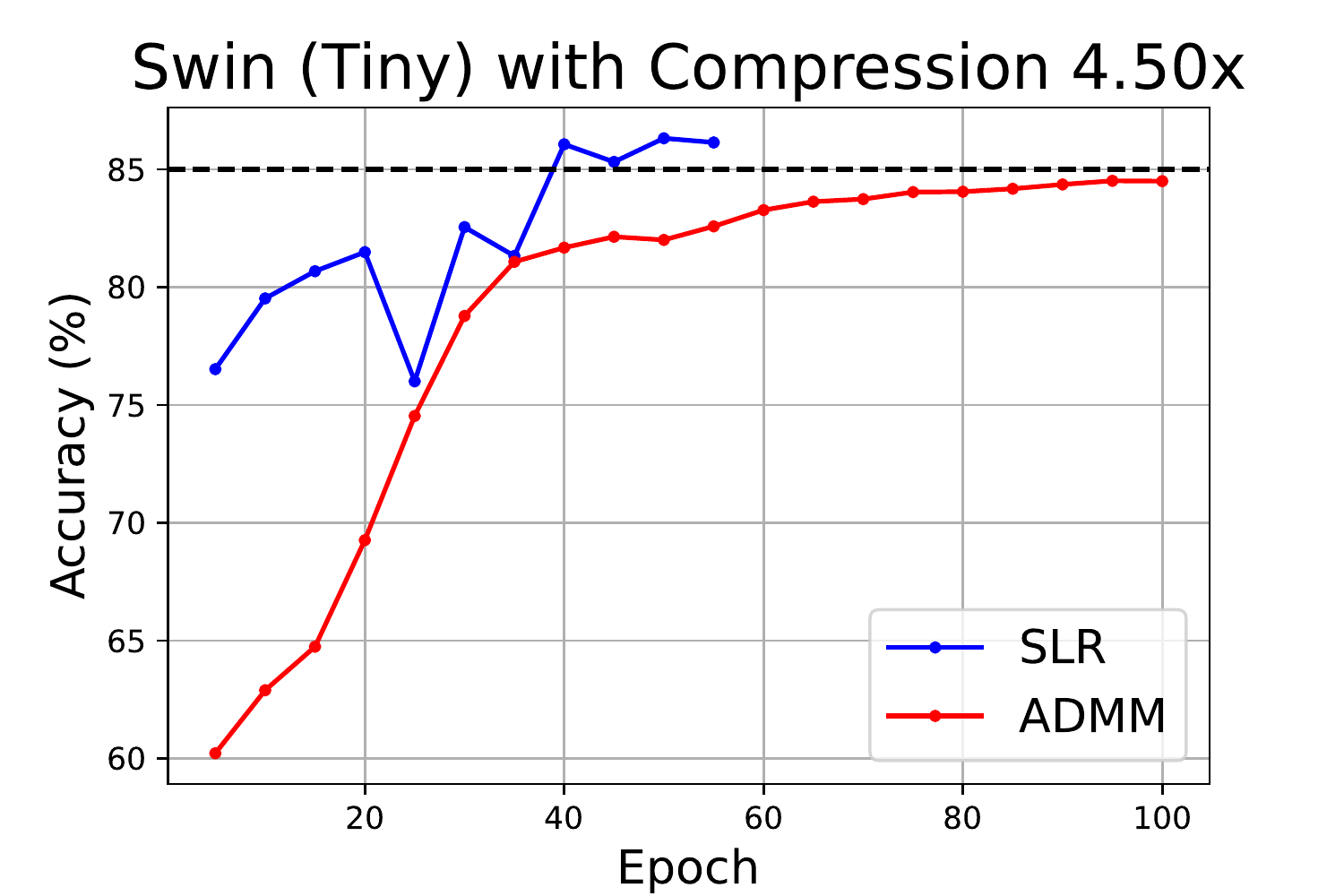}}
  \caption{Top-5 Hardprune accuracy of Swin Transformer (Tiny) on ImageNet benchmarks after applying SLR and ADMM weight pruning. Accuracy is reported periodically.}
\label{fig:hardprune_swin}
\end{figure*}

Figure~\ref{fig:hardprune_mlp} plots periodically checked the hardpruning accuracy of SLR and ADMM with MLP-Mixer on ImageNet. The compression rates correspond to Table~\ref{table:mlp}, following the same procedure described in section~\ref{classification}. As shown, SLR quickly converges and achieves higher accuracy compared with ADMM. When the compression rate is low, such as $1.96\times$, two methods converge at a similar rate. However, as the compression rate increases, SLR not only converges faster than ADMM but also reaches much higher accuracy. ADMM even cannot achieve the desired accuracy after $60$ epochs of training when the compression rate is $8.28\times$. In Figure ~\ref{fig:mlp_0.9}, we can see SLR quickly converges to the threshold at around $50$ epochs and results in an $8.6\%$ accuracy gap between ADMM when the training of SLR is stopped, ADMM's accuracy continues to improve albeit slowly until epoch $100$.

Figure~\ref{fig:hardprune_swin} plots periodically-checked hardpruning accuracy of SLR and ADMM with Swin Transformer on ImageNet, with the compression rates corresponding to those reported in Table~\ref{table:swin}. Under the three compression rates, SLR consistently converges to the desired accuracy in half the number of training epochs compared to ADMM. Even when the compression rate is $3.16\times$, SLR results in an approximately $1\%$ accuracy gap with ADMM when it stops training. This further demonstrates our SLR can quickly recover the model accuracy at the hard-pruning stage itself.


\subsection{Evaluation on Object Detection and Segmentation Tasks}
\label{object}

In this subsection, we evaluate our SLR-based weight pruning method on three object detection and segmentation benchmarks.

\paragraph{\textbf{Models and Datasets.}} In the first experiment, we test YOLOv3 and YOLOv3-tiny models \cite{yolov3} on COCO 2014 benchmark. We followed the publicly available Ultralytics repository\footnote{\url{https://github.com/ultralytics/yolov3}} for YOLOv3 and its pretrained models. The second experiment focuses on lane detection. We use the pretrained model from Ultra-Fast-Lane-Detection~\cite{qin2020ultra} on the TuSimple lane detection benchmark dataset. The third experiment involves 3D point cloud object detection experiments. We use PointPillars~\cite{lang2019pointpillars} pretrained model on KITTI 2017 dataset following the OpenPCDet repository\footnote{\url{https://github.com/open-mmlab/OpenPCDet}}.
In this experiment, we use LIDAR point cloud data as input. Each point cloud data point is stored as a large collection of 3D elevation points, each point is represented as a $1*4$ vector containing $x,y,z$ (3D coordinates) and intensity~\cite{zhou2021end}. 

\paragraph{\textbf{Training Settings.}} In all experiments we use $\rho = 0.1$. We set SLR parameters as $M = 300$, $r = 0.1$ and $s_0 = 10^{-2}$. We follow the same training settings provided by the repositories we use. Finally, we use the same number of training epochs and sparsity configurations for ADMM and SLR.

\begin{table}[!hb]
\parbox{.43\linewidth}{
\caption{ADMM and SLR pruning results with different structure of YOLOv3 on COCO dataset.}
\label{table:addmvsslr_coco}
\begin{tabular}{ccccc}
\hline\hline
\textbf{Architecture}                               
& \textbf{Epoch}      
& \textbf{Method} 
& \textbf{\begin{tabular}[c]{@{}c@{}}Hardprune \\ mAP\end{tabular}}
& \textbf{\begin{tabular}[c]{@{}c@{}}Comp. \\ Rate\end{tabular}} \\ 
\hline
\multirow{2}{*}{}                                             
& \multirow{2}{*}{15} & ADMM & 35.2 & \multirow{2}{*}{1.19$\times$}\\
&                     & SLR  & \textit{36.1} &                     \\
\cline{2-5} 
\multirow{2}{*}{\begin{tabular}[c]{@{}c@{}}YOLOv3-tiny \\ \textit{(mAP = 37.1)}\end{tabular}} 
& \multirow{2}{*}{20} & ADMM & 32.2 & \multirow{2}{*}{2$\times$} \\
&                     & SLR  & \textit{36.0} &                   \\ 
\cline{2-5} 
\multirow{2}{*}{}                                             
& \multirow{2}{*}{25} & ADMM & 25.3 & \multirow{2}{*}{3.33$\times$} \\
&                     & SLR  & \textit{35.4} &                  \\ 
\hline
\multirow{2}{*}{\begin{tabular}[c]{@{}c@{}}YOLOv3-SPP \\ \textit{(mAP = 64.4)}\end{tabular}}  
& \multirow{2}{*}{15} & ADMM & 41.2 & \multirow{2}{*}{2$\times$} \\
&                     & SLR  & \textit{53.2} &                   \\ 
\hline\hline
\end{tabular}
}
\hfill
\parbox{.45\linewidth}{
\caption{SLR pruning results with ResNet-18 on TuSimple benchmark through different compression rates.}
\label{table:lane_seg}
\begin{tabular}{ccccc}
\hline\hline
\multirow{2}{*}{\textbf{\begin{tabular}[c]{@{}c@{}}Comp. \\ Rate\end{tabular}}}
& \multicolumn{2}{c}{\textbf{\begin{tabular}[c]{@{}c@{}}Hardprune Acc. (\%)\end{tabular}}} 
& \multicolumn{2}{c}{\textbf{\begin{tabular}[c]{@{}c@{}}Retrain Acc. (\%)\end{tabular}}} \\ 
\cline{2-5} 
& \textbf{ADMM}  & \multicolumn{1}{c|}{\textbf{{SLR}}} & \textbf{ADMM} & \textbf{SLR} \\ 
\hline
$1.82\times$ & 92.49 & \multicolumn{1}{c|}{94.64} & 94.28 & 94.63 \\ 
\cline{2-5} 
$2.54\times$ & 92.25 & \multicolumn{1}{c|}{94.56} & 94.04 & 94.93 \\
\cline{2-5} 
$4.21\times$ & 90.97 & \multicolumn{1}{c|}{94.66} & 94.18 & 94.68 \\
\cline{2-5} 
$12.10\times$ & 88.41 & \multicolumn{1}{c|}{94.51} & 94.45 & 94.7 \\
\cline{2-5} 
$16.85\times$ & 78.75 & \multicolumn{1}{c|}{94.55} & 94.23 & 94.65 \\
\cline{2-5} 
$22.80\times$ & 67.79 & \multicolumn{1}{c|}{94.62} & 94.08 & 94.55 \\
\cline{2-5} 
$35.25\times$ & 57.05 & \multicolumn{1}{c|}{93.93} & 93.63 & 94.34 \\
\cline{2-5} 
$77.67\times$ & 46.09 & \multicolumn{1}{c|}{89.72} & 88.33 & 90.18 \\
\hline\hline
\end{tabular}
}
\end{table}

\begin{figure}[!hb]
\centering
\begin{minipage}{0.45\linewidth}
  \centering
  \subcaptionbox{YOLOv3. \label{fig:yolo_hardprune}}{\includegraphics[width=0.48\linewidth]{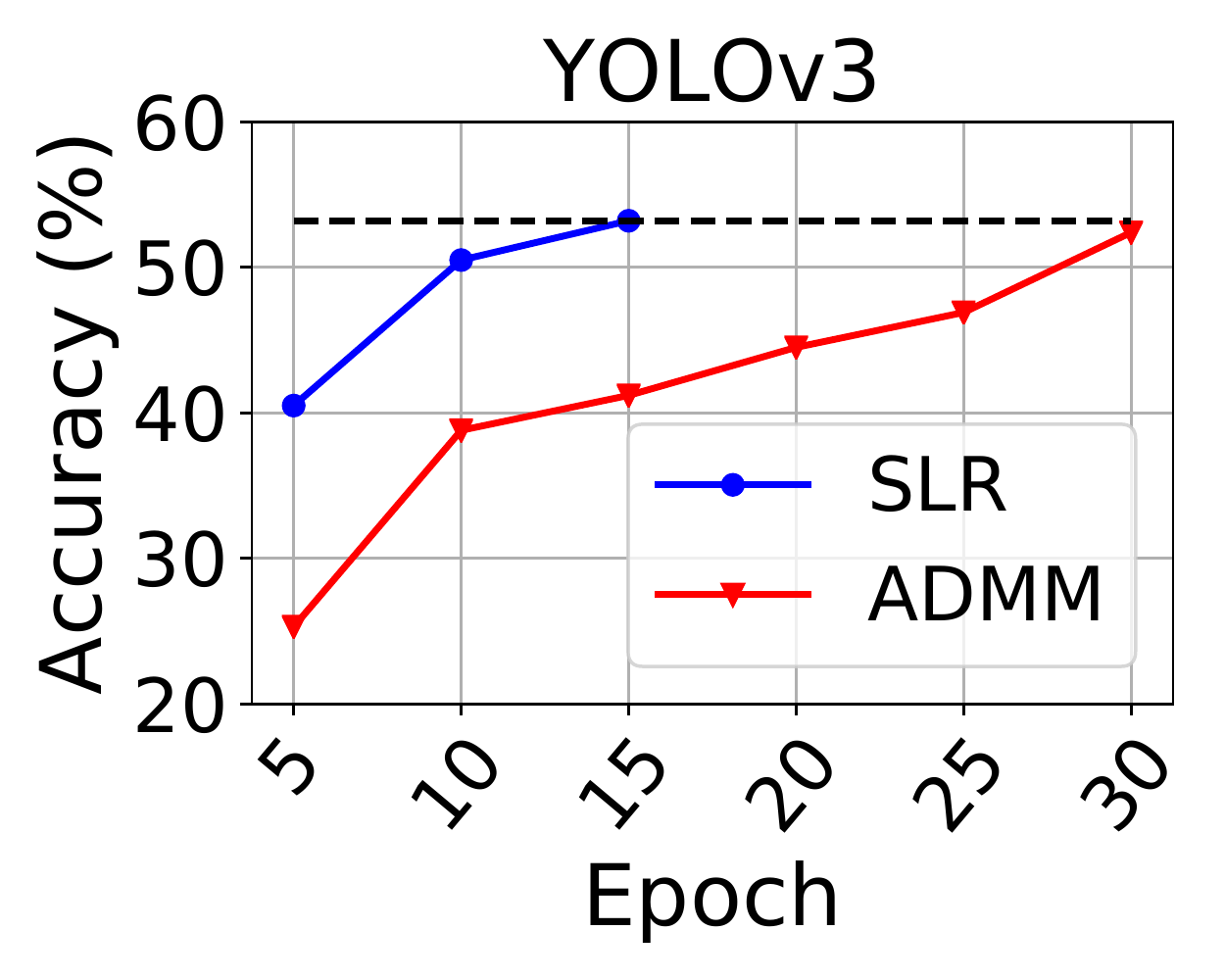}}
  \hspace{0em}%
  \subcaptionbox{YOLOv3-tiny. \label{fig:yolotiny_hardprune}}{\includegraphics[width=0.48\linewidth]{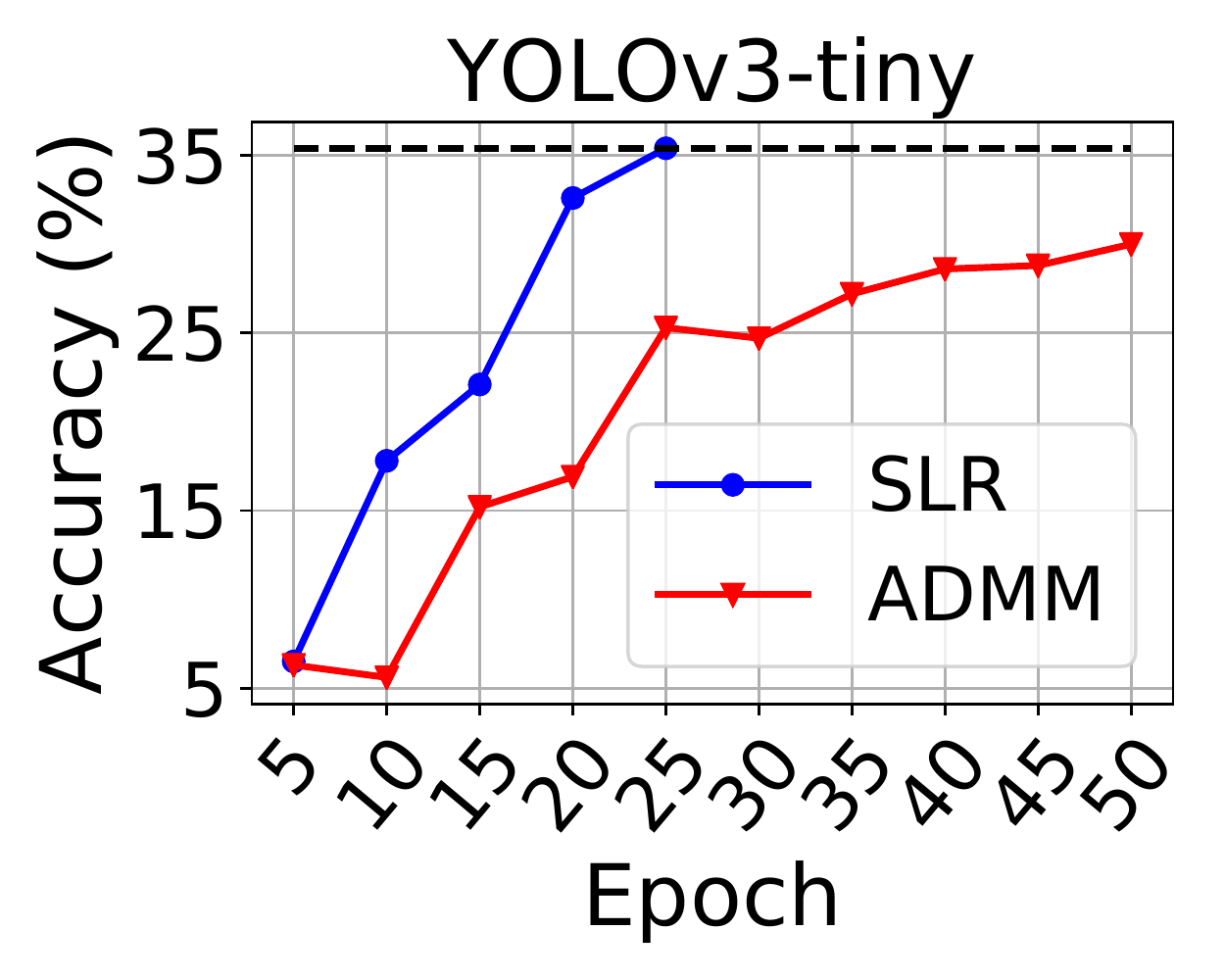}}
  \caption{Hardpruning accuracy of YOLOv3 and YOLOv3-tiny. Accuracy is reported every 5 epochs and training is stopped when methods reach the accuracy threshold.}
  \label{fig:hardprune_yolo}
\end{minipage}
\hspace{1em}
\begin{minipage}{0.45\linewidth}
  \centering
  \includegraphics[width=0.75\linewidth]{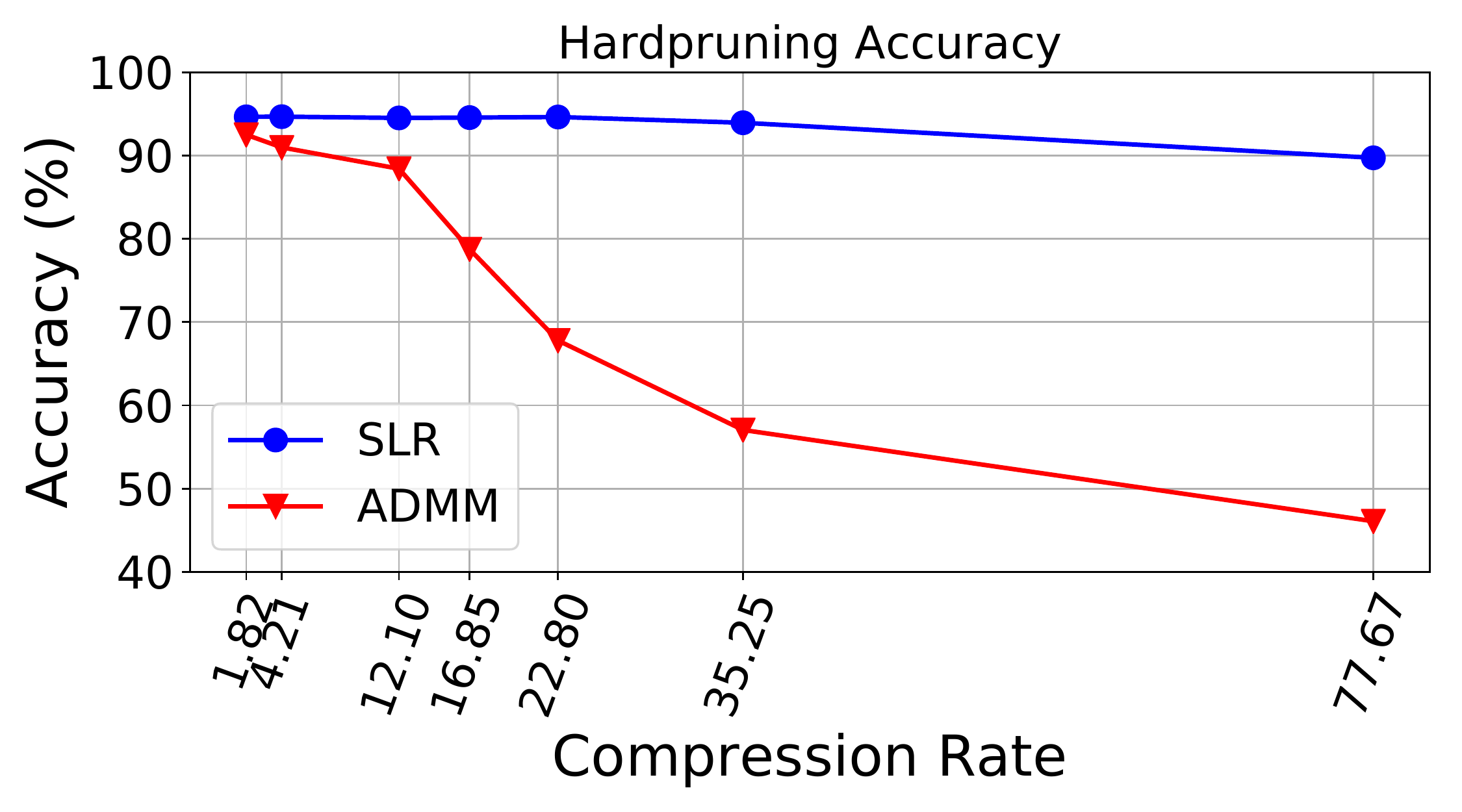}
  \captionof{figure}{Hardpruning accuracy on TuSimple benchmark with ADMM vs. SLR training for several compression rates. SLR has a greater advantage over ADMM as the compression rate increases.}
  \label{fig:tusimple_comp}
\end{minipage}
\end{figure}

\paragraph{\textbf{Testing Settings.}} On YOLOv3 models, we calculate the COCO mAP with IoU = 0.50 with image size of $640$ for testing. In lane detection experiments, the evaluation metric is ``accuracy", which is calculated as $\frac{\sum_{\text {clip}} C_{\text {clip}}}{\sum_{\text {clip}} S_{\text {clip}}}$, where $C_{clip}$ is the number of lane points predicted correctly and $S_{clip}$ is the total number of ground truth in each clip. 

The KITTI dataset is stratified into easy, moderate, and hard difficulty levels. Here, \textit{Easy level} means the minimum height of the bounding box is 40 pixels, all objects in the images are fully visible, and the percentage of truncation of objects being less than 15\%; \textit{Moderate level} is with minimum bounding box height being 25 pixels, objects in the images are partly visible, and percentage of truncation of objects being less than 30\%; \textit{Hard level} is with minimum bounding box height being 25 pixels, but some objects in the image are difficult to see, and maximum percentage of truncation is $50\%$. mAP is calculated under each difficulty strata with $IoU = 0.5$.

\paragraph{\textbf{Comparison of SLR and ADMM.}} Our comparison of SLR and ADMM methods on the COCO dataset is shown in Table \ref{table:addmvsslr_coco}. We have compared the two methods under three different compression rates for YOLOv3-tiny and tested YOLOv3-SPP pretrained model with a compression rate of $1.98\times$. We can see that the model pruned with the SLR method has higher accuracy after hardpruning in all cases. At a glance at YOLOv3-tiny results, we observe that the advantage of SLR is higher with an increased compression rate.

\begin{figure*}[!b]
  \centering
  \subcaptionbox{Weights before after pruning with SLR (middle) and ADMM (right) under the same compression rate (77.6$\times$). 
  \label{fig3:tusimple_a}}{\includegraphics[width=0.95\columnwidth]{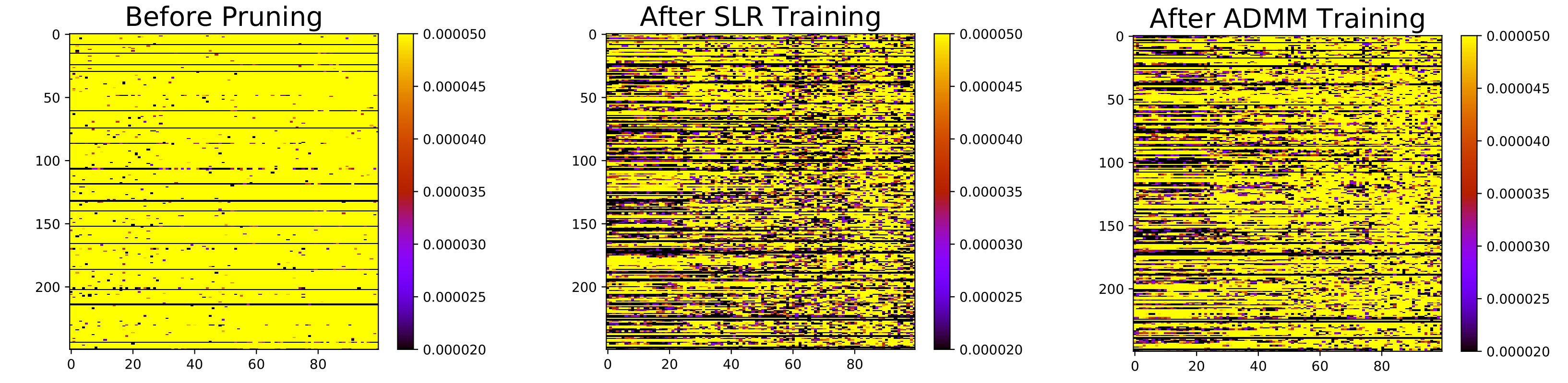}}
  \hspace{5em}%
  \subcaptionbox{Weights before and after pruning with SLR (middle) and ADMM (right) under the same accuracy (89.0\%). 
  \label{fig3:tusimple_b}}{\includegraphics[width=0.95\columnwidth]{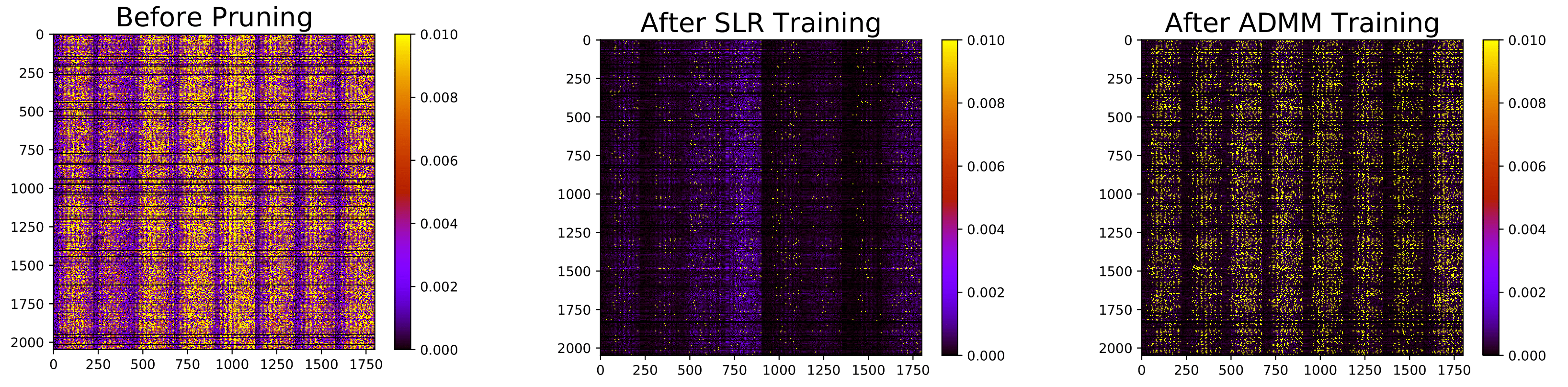}}
  \caption{Heatmap of ResNet-18 weights on TuSimple benchmark before and after pruned with SLR and ADMM. Weights are more zeroed out with SLR compared to ADMM. }
\label{fig:tusimple_heat}
\end{figure*}

In compression rate $3.33\times$ on YOLOv3-tiny, given a limit of $25$ epochs, we can observe that the gap between ADMM and SLR is much higher, which is due to the faster convergence of SLR as shown in Figure \ref{fig:yolotiny_hardprune}. Similarly, Figure \ref{fig:yolo_hardprune} shows the mAP progress of YOLOv3 during SLR and ADMM training for $50$ epochs, pruned with $2\times$ compression rate. SLR reaches the mAP threshold only at epoch $15$.

Table \ref{table:lane_seg} reports our result for the Lane Detection task on the TuSimple lane detection benchmark after $40$ epochs of training and $5$ epochs of masked-retraining. We conducted experiments under $8$ different compression rates. Figure \ref{fig:tusimple_comp} illustrates the accuracy gap between ADMM and SLR methods after hardpruning as the compression rate increases. 

From Figure \ref{fig:tusimple_comp}, our observation is that for a small compression rate such as $1.82\times$, SLR has little advantage over ADMM in terms of hardpruning accuracy. However, as the compression rate increases, SLR starts to perform better than ADMM. For example, SLR survives $77.67\times$ compression with slight accuracy degradation and results in $89.72\%$ accuracy after hardpruning, while ADMM accuracy drops to $46.09\%$. This demonstrates that our SLR-based training method has a greater advantage over ADMM, especially in higher compression rates, as it achieves compression with less accuracy loss and reduces the time required to retrain after hardpruning.

\begin{table}[!b]
\caption{ADMM and SLR results of PointPillars model on KITTI in different task difficulty levels and compression rates.}
\label{table:pointcloud_table}
\centering
\begin{tabular}{ccc|cc|cc}
\hline
\multirow{2}{*}{\textbf{Level}} & \multirow{2}{*}{\textbf{Orig mAP}} & \multirow{2}{*}{\textbf{Compression}} & \multicolumn{2}{c}{\textbf{ADMM}} & \multicolumn{2}{c}{\textbf{SLR}} \\ \cline{4-7} 
 & & & \textbf{After Hardprun} & \textbf{After Retrain} & \textbf{After Hardprune} & \textbf{After Retrain} \\ \hline
\multirow{4}{*}{Easy} & \multirow{4}{*}{80.7} & $4.874\times$ & 77.0 & 81.5 & 77.8 & 82.2 \\
 & & $5.702\times$ & 74.7 & 74.7 & 77.3 & 79.3 \\
 & & $6.431\times$ & 72.9 & 77.5 & 76.6 & 75.3 \\
 & & $9.449\times$ & 58.6 & 70.9 & 68.0 & 79.9 \\ \hline\hline
\multirow{4}{*}{Moderate} & \multirow{4}{*}{78.5} & $4.874\times$ & 73.8 & 77.1 & 74.5 & 78.4 \\
 & & $5.702\times$ & 71.9 & 71.0 & 74.5 & 75.3 \\
 & & $6.431\times$ & 69.9 & 73.2 & 73.2 & 72.7 \\
 & & $9.449\times$ & 54.1 & 66.8 & 65.6 & 75.8 \\ \hline\hline
\multirow{4}{*}{Hard} & \multirow{4}{*}{60.7} & $4.874\times$ & 51.9 & 51.2 & 52.0 & 56.4 \\
 & & $5.702\times$ & 50.1 & 50.2 & 52.9 & 51.3 \\
 & & $6.431\times$ & 48.5 & 47.8 & 49.5 & 50.7 \\
 & & $9.449\times$ & 30.8 & 35.6 & 46.1 & 51.4 \\ \hline
\end{tabular}
\end{table}

\begin{figure*}[!b]
  \centering
  \subcaptionbox{Pruning on Easy task level. \label{fig3:a}}{\includegraphics[width=0.3\columnwidth]{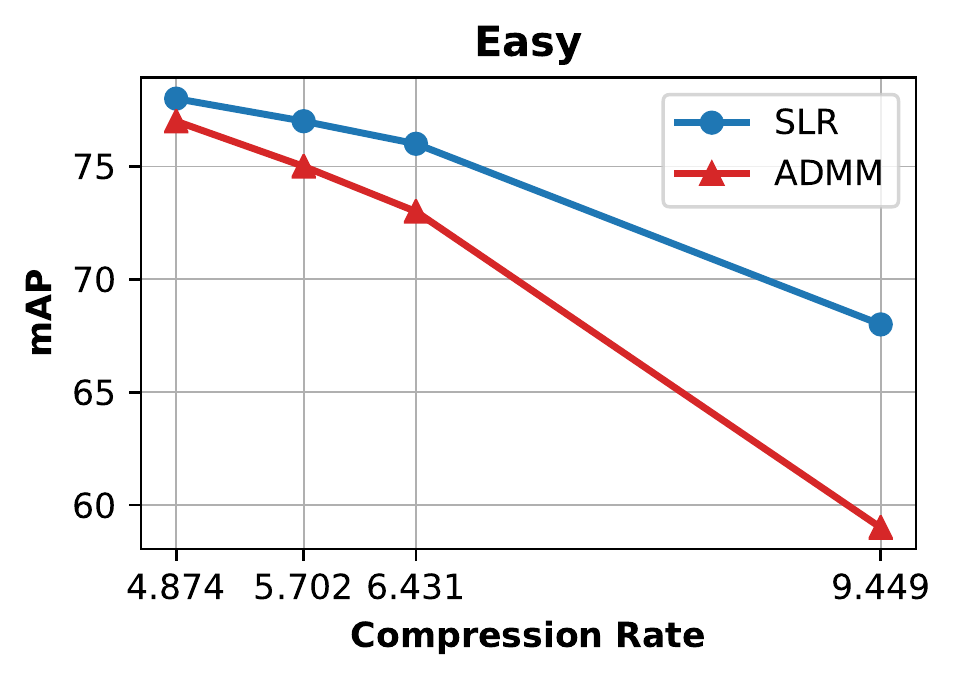}}\hspace{1em}%
  \subcaptionbox{Pruning on Moderate task level.\label{fig3:b}}{\includegraphics[width=0.3\columnwidth]{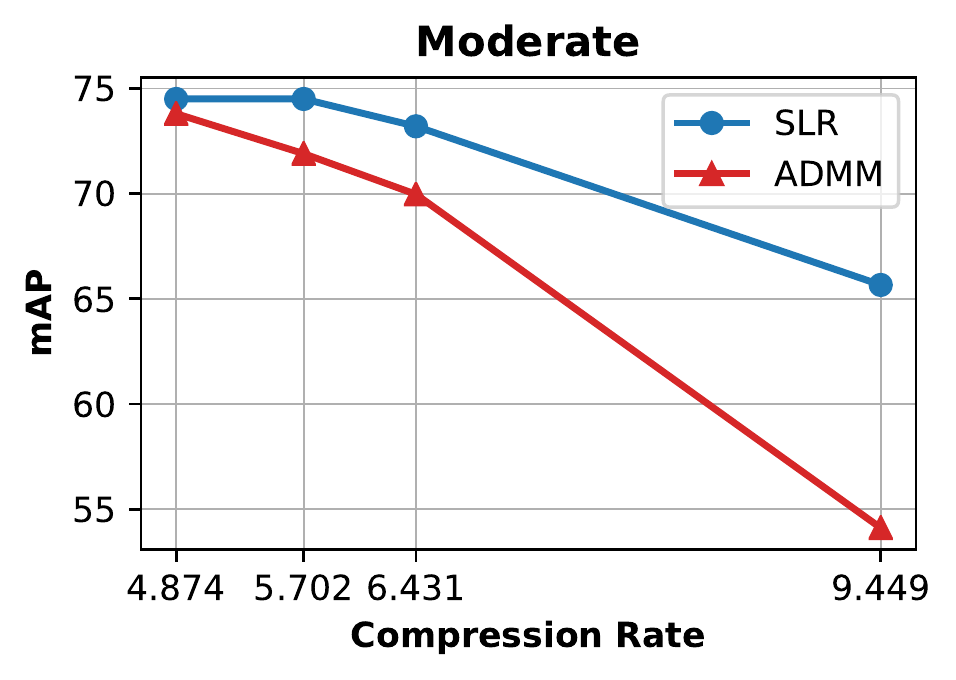}}\hspace{1em}%
  \subcaptionbox{Pruning on Hard task level.\label{fig3:c}}{\includegraphics[width=0.3\columnwidth]{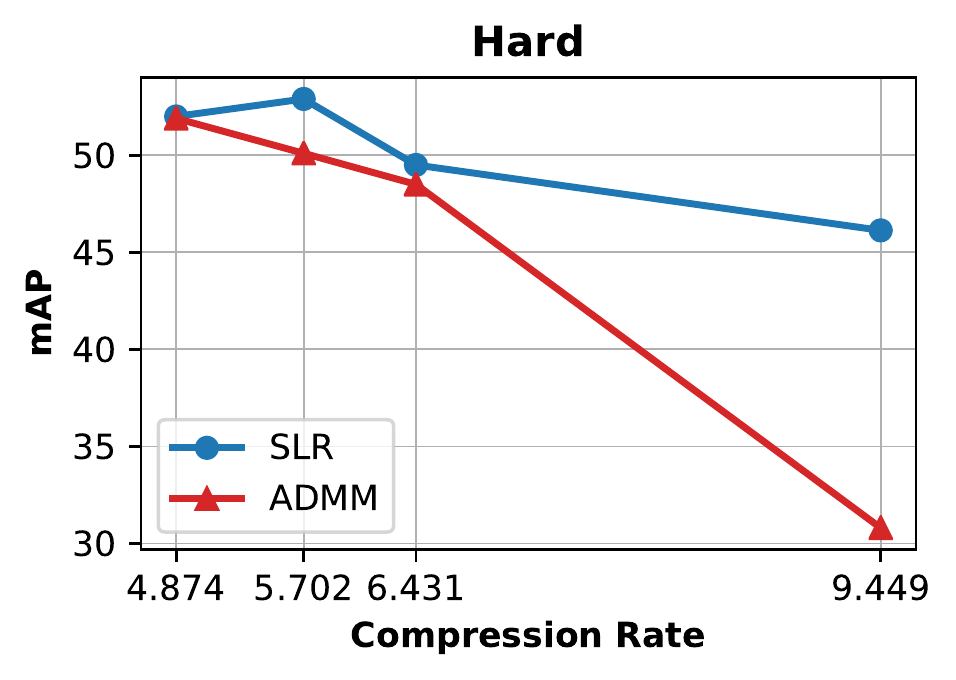}}
  \caption{mAP of PointPillars model after hard-pruning using different compression rates under different difficulty levels.}
\label{fig:pointcloud_fig}
\end{figure*}

Finally, in Figure \ref{fig:tusimple_heat}, we show the difference between the weights of one layer before and after pruning with SLR and ADMM. In Figure \ref{fig3:tusimple_a}, we show the initial (non-pruned) weights and then show the sparsity of weights under the same compression rate ($77\times$) with SLR and ADMM. Initially, the layer has low sparsity. After training with SLR and ADMM, we can see an increased number of zeroed-weights. SLR moves towards the desired sparsity level faster than ADMM. In Figure \ref{fig3:tusimple_b}, we compare the sparsity of weights under the same accuracy ($89.0\%$). It can be observed that SLR significantly reduced the number of non-zero weights, and ADMM has more non-zero weights remaining compared with SLR.

Lastly, Table \ref{table:pointcloud_table} shows the 3D point cloud object detection model compression results on the KITTI benchmark. The SLR and ADMM results are reported after $40$ epochs of training and $3$ epochs of masked-retraining. We have two observations: First, SLR has much higher accuracy than ADMM after hardpruning with compression rate increased. For example, as shown in Figure \ref{fig:pointcloud_fig}, when tested under $9.44\times$ compression rate, under ``hard" strata, SLR has mAP that more than $15\%$ higher than ADMM as illustrated in Figure \ref{fig3:c}. In ``easy" and ``moderate" difficulty strata, as shown in Figure \ref{fig3:a} and Figure \ref{fig3:b}, hardpruning mAP of SLR is still more than $10\%$ higher than ADMM under $9.44\times$ compression rate. Secondly, we observe that since SLR has higher mAP after the hardpruning stage, it also reaches significantly higher mAP after retraining. There is almost $16\%$ mAP difference between SLR and ADMM in the ``hard" difficulty level under $9.44\times$ compression rate. This shows that when the retraining budget is limited, our SLR method can quickly recover the model accuracy.

\subsection{Ablation Studies}
\label{ablation}

\begin{figure}[!ht]
  \centering
  \subcaptionbox{$s_0$ parameter. \label{fig3:finetuning_s}}{\includegraphics[width=0.4\columnwidth]{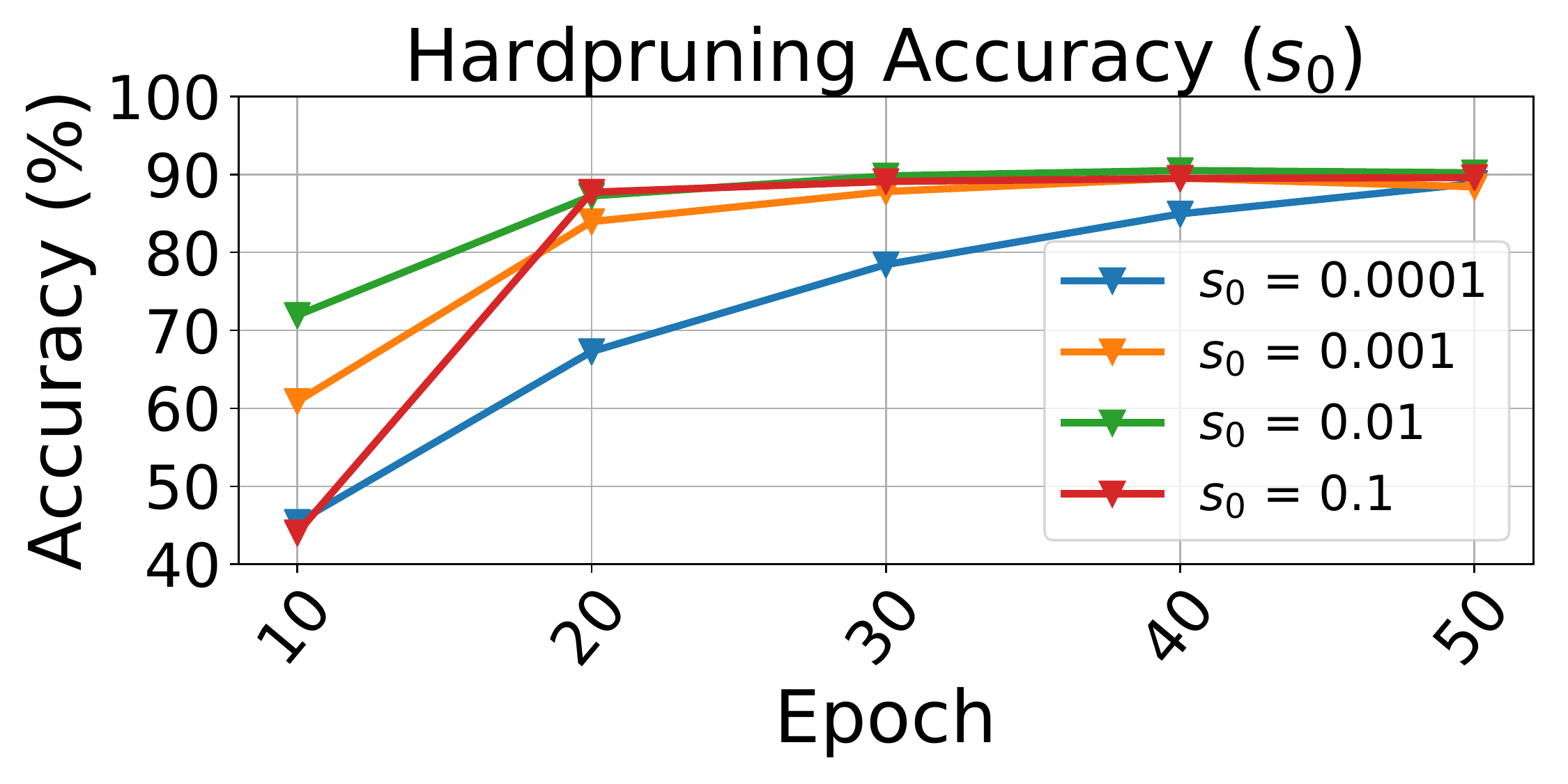}}\hspace{1em}%
  \subcaptionbox{$M$ parameter. \label{fig3:finetuning_m}}{\includegraphics[width=0.2\columnwidth]{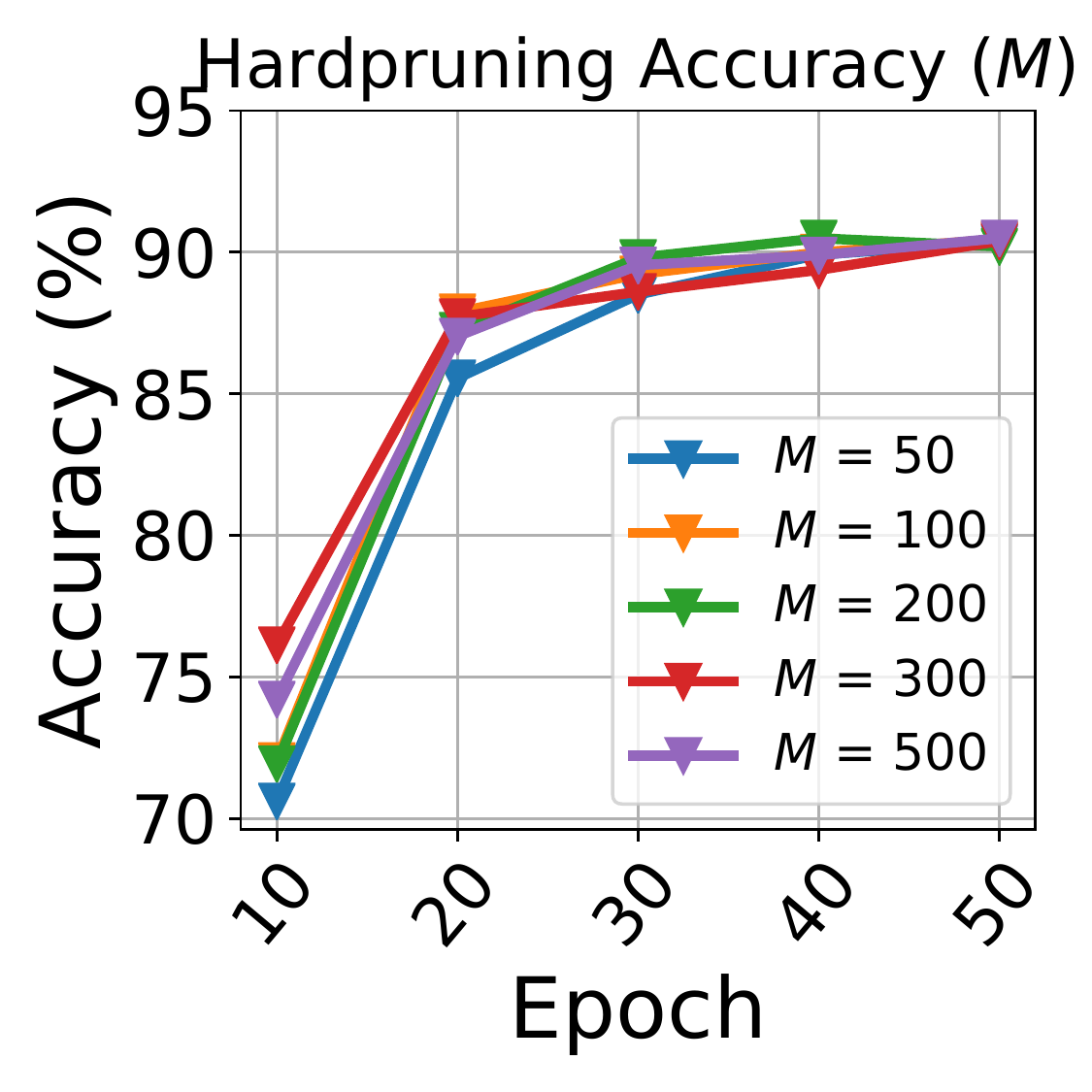}}\hspace{1em}%
  \subcaptionbox{$r$ parameter.\label{fig3:finetuning_r}}{\includegraphics[width=0.2\columnwidth]{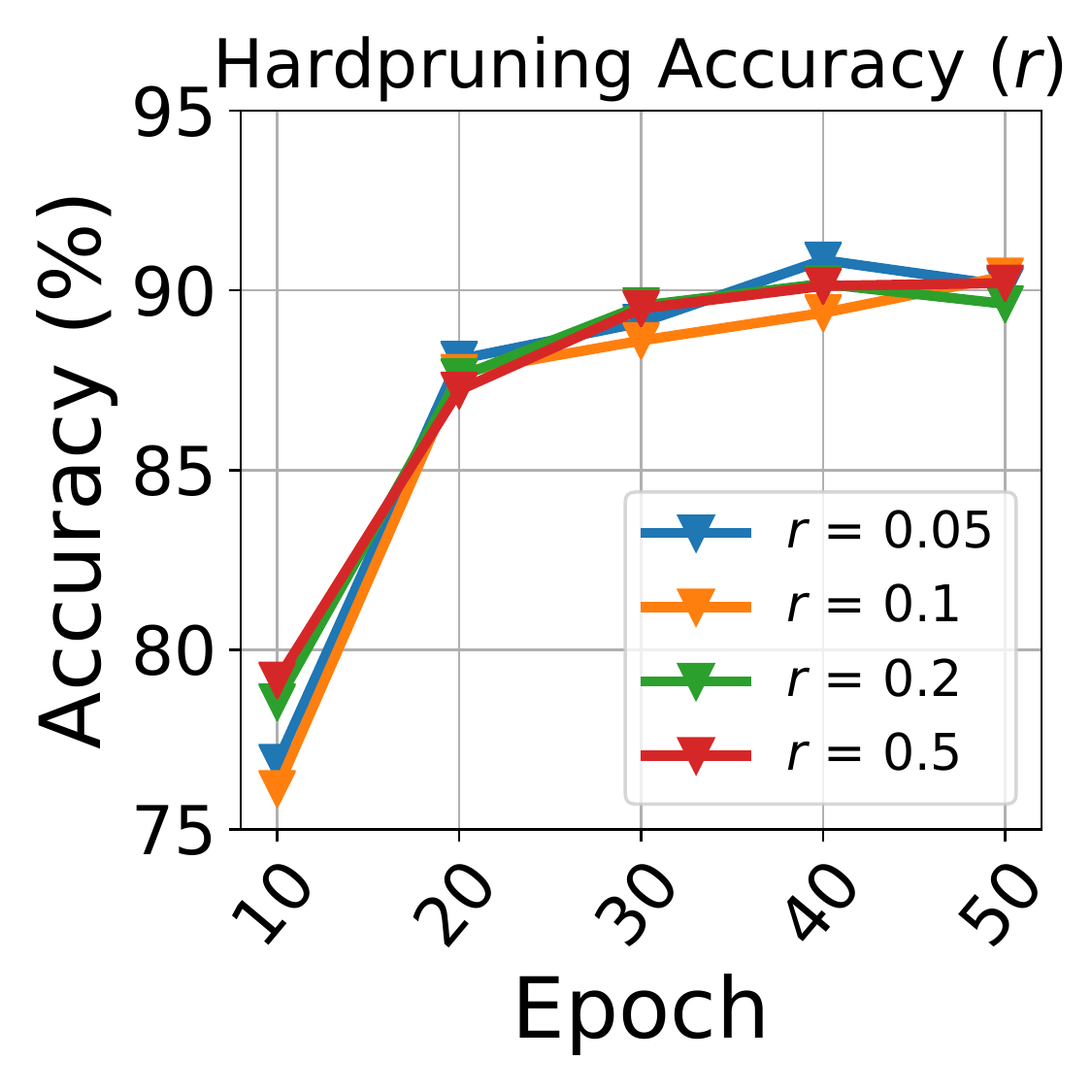}}
  \caption{Hardpruning accuracy of ResNet-18 on CIFAR-10 during SLR training with respect to different values of $s_0$, $M$ and $r$.}
\label{fig:finetuning_figs}
\end{figure}
\begin{wrapfigure}{l}{0.5\textwidth}
  \begin{center}
    \includegraphics[width=0.48\textwidth]{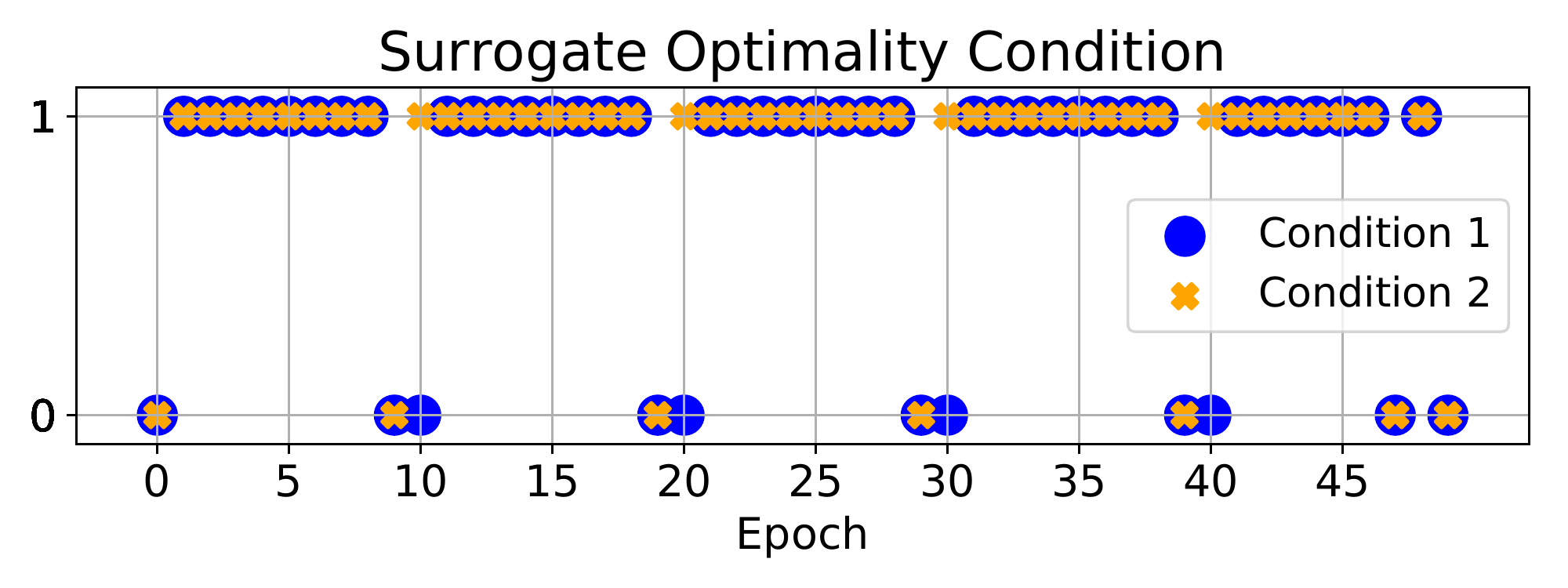}
  \end{center}
  \caption{Surrogate optimality condition satisfaction graph during the SLR training of ResNet-18 on CIFAR-10 for 50 epochs (1: satisfied, 0: not satisfied). Conditions are satisfied periodically.}
  \label{fig:opt_condition_50epochs}
\end{wrapfigure}

We conduct several experiments to observe SLR behavior with respect to SLR parameters $\rho, s_0, r$ and $M$ on the ResNet-18 model ($93.33\%$ accuracy) on CIFAR-10. We prune the model through SLR training for $50$ epochs with a compression rate of $8.71\times$ and observed the hardpruning accuracy every $10$ epochs. Figure \ref{fig:finetuning_figs} shows the accuracy of the model through SLR training based on the different values of $s_0$, $M$, and $r$. Based on the hardpruning accuracy throughout training, it can be seen that, even though the parameters do not have a great impact on the end result, the choice of $s_0$ can impact the convergence of the model. From Figure \ref{fig3:finetuning_s}, we can state that $s_0 = 10^{-2}$ provides higher starting accuracy and converges quickly. Figure \ref{fig3:finetuning_m} and Figure \ref{fig3:finetuning_r} demonstrate the impact of $M$ and $r$ on the hardpruning accuracy respectively.

Figure \ref{fig:opt_condition_50epochs} demonstrates that there exists iteration $\kappa$ (as required in the Theorem) so that the surrogate optimality condition, the high-level convergence criterion of the SLR method, is satisfied during training with $s_0 = 10^{-2}$, $\rho=0.1$ thereby signifying that ``good" multiplier-updating directions are always found. For example, after the conditions are violated at epoch $9$, there exits $\kappa = 10$ so that at iteration $11$, after $\kappa = 10$, the surrogate conditions are satisfied again. 



\section{Conclusions}

In this paper, we addressed the DNN weight-pruning problem as a non-convex optimization problem by adopting the cardinality function to induce weight sparsity. The SLR method decomposes the relaxed weight-pruning problem into subproblems, which are then efficiently coordinated by updating Lagrangian multipliers, resulting in fast convergence. We carried out weight-pruning experiments on image classification and object detection and segmentation tasks on various datasets to compare our SLR method against ADMM and other SoTA. 
We observed that our SLR method offers a significant advantage under high compression rates and achieves higher accuracy during weight pruning. Additionally, SLR reduces the accuracy loss caused by the hardpruning and shortens the retraining process. Given its effective optimization as well as coordination capabilities and clear advantages demonstrated through various examples, the SLR method holds strong potential for broader DNN-training applications.  



\bibliographystyle{ACM-Reference-Format}
\bibliography{sample-base}




\end{document}